\documentclass{article}

\usepackage{PRIMEarxiv}

\usepackage[utf8]{inputenc} % allow utf-8 input
\usepackage[T1]{fontenc}    % use 8-bit T1 fonts
\usepackage{hyperref}       % hyperlinks
\usepackage{url}            % simple URL typesetting
\usepackage{booktabs}       % professional-quality tables
\usepackage{amsfonts}       % blackboard math symbols
\usepackage{nicefrac}       % compact symbols for 1/2, etc.
\usepackage{microtype}      % microtypography
\usepackage{lipsum}
\usepackage{fancyhdr}       % header
\usepackage{graphicx}       % graphics
\usepackage{subcaption} % Subcaptions
\graphicspath{{media/}}     % organize your images and other figures under media/ folder

\title{Deep Learning Improvements For Sparse \protect\\ Spatial Field Reconstruction}
\author{Robert Sunderhaft$^1$ \hspace{1.5cm} Logan Frank$^2$ \hspace{1.5cm} Jim Davis$^2$ \\ \\
$^1$\ Department of Data Analytics \\
$^2$\ Department of Computer Science and Engineering \\
Ohio State University \\
{\tt\small \{\ sunderhaft.2, frank.580, davis.1719\ \}@osu.edu}
}

\begin{document}
\maketitle

\begin{abstract}
%  The dissertation abstract can only be 500 words.

Accurately reconstructing a global spatial field from sparse data has been a longstanding problem in several domains, such as Earth Sciences and Fluid Dynamics. Historically, scientists have approached this problem by employing complex physics models to reconstruct the spatial fields. However, these methods are often computationally intensive. With the increase in popularity of machine learning (ML), several researchers have applied ML to the spatial field reconstruction task and observed improvements in computational efficiency. One such method \cite{Fukami2021} utilizes a sparse mask of sensor locations and a Voronoi tessellation with sensor measurements as inputs to a convolutional neural network for reconstructing the global spatial field. In this work, we propose multiple adjustments to the aforementioned approach and show improvements on geoscience and fluid dynamics simulation datasets. We identify and discuss scenarios that benefit the most using the proposed ML-based spatial field reconstruction approach.

\end{abstract}

% keywords can be removed

% Introduction Section
\section{Introduction}
\label{intro.ch}

The world lives in the information era, a time period characterized by new technology, rapid communication, and an ever growing corpus of data \cite{Castells1997}. Before 2003, humans generated a cumulative  $10^{18}$ bytes of data, but in 2013, we created that amount of information in only two days \cite{Sagiroglu2013}. While there is an unprecedented amount of complete and well-curated data today, there still exists domains and problems where obtaining complete data is infeasible, limiting a user to only sparse information. One such example is geoscience, where it is crucial to have complete global representations of physical systems to predict weather, climate change, natural disasters, and more. However, not all physical measurements can be estimated at a global scale. Many problems are limited to specific data sensors like subsurface ocean temperatures and salinity. Furthermore, for problems where we do have widespread global observations, we lack dense representations from before the 1980s, causing geoscience models to ignore valuable historical data. Similarly, in fluid dynamics it is infeasible to measure variables at every possible location, thus making full fluid flow reconstruction from a sparse amount of sensors an important problem. Having the ability to accurately reconstruct complete spatial data from a sparse amount of sensors can provide many benefits, including reconstructing historical climate datasets to improve climate change models, achieving better active flow control in fluid dynamics, and decreasing the latency of using reanalysis models.

% By accurately reconstructing fluid flows from a sparse amount of sensors, we can achieve better active flow control and potentially help craft more fuel-efficient automobiles and high-efficiency turbines. (Finish citing)

Historically, disciplines that need to obtain global representations from sparse data have relied on physically informed models \cite{Carrassi2018}, simulations \cite{Solman2013}, and solving differential equation models \cite{Liu1996}, all of which are computationally expensive. For example, in fluid dynamics the Navier-Stokes equations are used to describe fluid motion near perfectly, but its computational complexity is so great that it cannot be employed to solve problems regarding weather or climate \cite{Rowley2017}. Recently, researchers have looked at machine learning (ML) as a potential two-fold solution for both lowering computational cost of spatial reconstruction and replacing the aforementioned simulations all together. One particular method that applies ML to the task of global field reconstruction for decreasing computational cost is \cite{Fukami2021} where they employ a convolutional neural network (CNN) to reconstruct the spatial field from two input images: a sparse mask image of the sensor locations and a Voronoi tessellation of the sensor measurements. While this approach showed good performance and was able to handle the potential issues surrounding moving and variable amounts of sensors, it exhibits many areas for potential improvement.

In this paper, we propose adjustments to the approach of \cite{Fukami2021} by altering the input data used for training. In particular, we 1) compute and use the Distance Transform on the sparse sensor location mask, increasing the amount of location information provided to the network, 2) normalize the sensor location mask, the Voronoi diagram, and the ground truth reconstruction to have values in the range [-1, 1], and 3) show for a sea surface temperature dataset that performance gains can be obtained by decoupling the external land information from the Voronoi diagram, providing insights on how domain knowledge can be correctly incorporated into ML-based spatial field reconstruction methods. Finally, we use our developed methodology to reconstruct a fourth dataset, Antarctic Surface Temperature, not explored by \cite{Fukami2021} to demonstrate the continued effectiveness of our method.

\subsection{Contributions and Significance}

This paper introduces novel data augmentations that can be used to improve sparse spatial field reconstruction. Specifically:

\begin{itemize}
  \item \textbf{The Distance Transform Mask:} An input image to a Deep Learning model that denotes the distance of a location to the nearest sensor. This input image introduces information regarding the spatial relationship between points in a spatial field allowing the Deep Learning model to understand that locations far away from sensors have less information than locations close to sensors.
  
  \item \textbf{The Land Mask:} An input image that denotes the locations in a map that are meant to be masked out or have no real values in the reconstructed image. Previous methods combined the Land Mask with other input data, which in turn introduces ambiguity into the data. 

  \item \textbf{Image Normalization:} A simple normalization that bounds the values of all input images, the Distance Transform Mask, Voronoi Mask, Ground Truth, etc, to a range of [-1,1]. Previous methods used no normalization, resulting in slower reconstruction speeds.
  
\end{itemize}

 Utilizing, the above data augmentations resulted in a 2x to 6x increase in training speed compared to previous Deep Learning based reconstruction methods as well as a decrease in reconstruction error.

 By improving sparse spatial field reconstruction, we are presenting a method that geoscientists can use to accurately reconstruct historical data or reduce the latency of current reanalysis models. This increase of  available data is useful to understand the effects of climate change both historically and in the future. Furthermore, the field of fluid dynamics will benefit from more accurate reconstructions by being able to make better decisions, like active flow control, under limited information.

\subsection{Organization}

This paper is organized as follows. We begin with a review of existing literature for sparse spatial field reconstruction and evaluate how these current methods can be improved upon in Chapter 2. Chapter 3 will introduce our proposed methodology to improving sparse spatial field reconstruction and detail the thought process for each improvement. Chapter 4 introduces the three types of datasets we used for experimentation. Chapter 5 provides an overview and analysis of the experimentation results followed by an application of our methodology to a new dataset in Chapter 6. The final chapter will provide a summary of the research, the significance, potential future work, and brief conclusion.

% Related Work Section
\section{Related Work}
\label{related_work.ch}

In recent years, many works have been proposed in the area of ML-based global field reconstruction. These approaches can be largely divided into three subcategories: 1) na\"{i}ve deep learning, 2) physics-informed deep learning, and 3) image in-painting.

\subsection{Na\"{i}ve Deep Learning}

Neural networks (NNs) are powerful ML models frequently employed for a variety of tasks including classification, regression, and more. In particular, CNNs are commonly used in the vision domain for recognizing, detecting, and/or segmenting classes in images and videos. Furthermore, NNs have been utilized in the image reconstruction task specifically applied to spatial field data. Such approaches often avoid using extremely sparse inputs because of the challenges they present, and when sparse inputs are used they only contain the exact sensor locations and the sensor values and do not account for the spatial relationship between sensors.

Many methods have na\"{i}vely applied NNs in geoscience for reconstructing global spatial fields of ocean subsurface temperature, water salinity, and other weather measurements \cite{Tian2022, Su2022, Su2021, Mao2023, Chen2022, Santos2022}. In \cite{Mao2023}, they attempt to reconstruct the ocean subsurface temperature and salinity using only the dense temperature information provided on the sea surface and disregard the sparse data from subsurface sensors. In \cite{Chen2022}, they do incorporate sparse sensor measurements of subsurface variables, but rather than transforming the sparse inputs into dense representations, they input only the sparse information.

The aforementioned methods utilize only the sparse inputs or none at all. In \cite{Fukami2021}, they propose an improvement to the typical reconstruction approach by separating the sparse sensor location information from the sensor values into another image representation called the Voronoi tessellation. The Voronoi is a spatially optimal projection of local sensor measurements onto the spatial domain, creating a dense representation for sensor values. They provide this Voronoi image and the sparse sensor location mask images as input to their CNN model which then outputs the reconstructed spatial field. By creating a Voronoi image, their method is able to create a less-sparse representation for the sensor values as the network now has information about the value at the closest sensor for each pixel in the image. However, their sensor location mask still remains sparse, which we found to be sub-optimal. We propose to pre-process the sensor location information further to create a dense representation of sensor locations. As we are proposing improvements to \cite{Fukami2021}, we will compare our work directly with theirs in our experiments.

\subsection{Physics-Informed Neural Networks}

Physics-informed neural networks (PINNs) are neural models that utilize physical equations in their loss functions during training \cite{Raissi2019}.
This enables the network to maintain and adhere to certain desirable laws of physics while leveraging the power of deep learning. PINNs have been widely adopted in physics \cite{Cail2020}, with particular interest in fluid dynamics for compressible flow \cite{Mao2020}, biomedical flow \cite{Yin2021}, turbulent convection flow \cite{Lucor2021}, and cylindrical flow \cite{Xu2022} problems. Our work does not utilize any equations of physical properties, however we will examine a subset of these flow problems in our experiments. Future iterations of this work could potentially benefit from adding physically informed loss functions for training our reconstruction models.

\subsection{In-painting}

The image in-painting task aims to restore missing or corrupted values in an image. When applied to the geophysics domain, it can be used to ``fill in'' regions of data that were not originally collected (i.e., spatial field reconstruction). Current research has used image in-painting to reconstruct the missing climate information in the Antarctic \cite{Yao2023}, missing historical climate information \cite{Kadow2020}, and sea surface temperatures in locations where clouds interfere with satellite measurements \cite{Dong2019}. While image in-painting may be optimal in situations where the data is mostly complete with only a few values missing, such as the cloud interference example, it is not applicable when there is a significant amount of missing data, such as reconstructing a 300x300 spatial field using only 50 sensor values. Thus, image in-painting is not general for all situations and in settings where there is a significant lack of data.

\subsection{Meta-Analysis}

The prior related-works mainly focus on modifying architecture of models, as in Na\"{i}ve Deep Learning and Image In-painting, or creating special loss functions, as in PINNs, but few if any focus on the input data. The goal of this work is to study the effects of varying data augmentations on inputs used in sparse spatial field reconstruction, which has the potential to improve upon all prior related-works.

% Methodology Section
\section{Methodology}
\label{methods.ch}

In this section, we describe the methods used in our image reconstructions and highlight our modifications from the original paper \cite{Fukami2021}: 1) applying the Distance Transform to the sensor location mask, 2) normalizing the input data, and 3) correctly utilizing domain knowledge.

The method proposed in \cite{Fukami2021} utilized two input images in a CNN to reconstruct the spatial field: a sparse (sensor) location mask image and a Voronoi tessellation generated from that sensor location mask, shown in Fig.~\ref{fig:sparse_loc_mask} and Fig.~\ref{fig:masked_voronoi}, respectively. The original paper’s sensor location mask was a binary image that displayed a 1 in pixels where sensors are located and 0 elsewhere, effectively creating a sparse representation of the sensor locations. The second image, the Voronoi tessellation, contains data about the sensor measurements. To obtain the corresponding Voronoi image, the closest sensor for every pixel in the image is found and set with the measurement of that closest sensor. The Voronoi diagram segments the spatial field into $n$ regions, where $n$ is the number of sensors. If an area exists where no measured values can be obtained, such as land masses in a sea surface temperature dataset, then the values corresponding to this area will be masked out with 0 in the Voronoi diagram. The sensor location and Voronoi tessellation images are the only inputs into the CNN used to reconstruct the complete spatial field in \cite{Fukami2021}.

\subsection{Voronoi}

\begin{figure}
    \centering
    \includegraphics[width = 5cm]{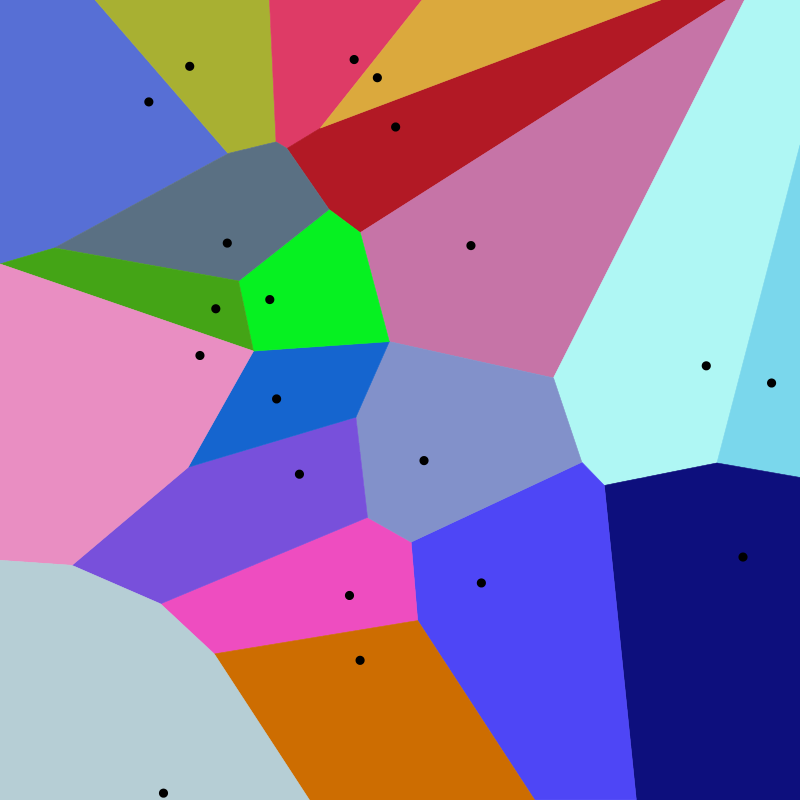}
    \caption{Voronoi Tessellation Example}
    \label{fig:ex_vor_tes}
\end{figure}

The Voronoi tessellation is an algorithm that subdivides $n$ key points on a plane into $n$ distinct regions. Each region denotes a set of points that are closest to a specific key point. This can also be understood as the visualization of a nearest-neighbor classifier where each distinct region is the set of nearest-neighbors to a given key point. An example of a Voronoi Tessellation can be seen in Fig.~\ref{fig:ex_vor_tes}. In this example there are 20 points on the plane divided into 20 different colored regions. All the points within a colored region are closest to the key point, represented by the black dot, within that region. Boundaries between regions denote points that are equidistant between two key points.

The Voronoi tessellation was used in our algorithm to segment the sparse spatial key points, sensors in our real world case, in our image to give an approximated value to points that were devoid of sensors. In this case, the regions do not represent categories, but instead is tied to a specific numerical measurement representing the best approximation for the general region. In the case of the NOAA dataset, which tracks the sea surface temperatures of the ocean, if we were given temperature measurements from a limited amount of temperature sensors we could use the Voronoi tessellation algorithm to give the entire earth an approximation of sea surface temperature, which can be visualized in Fig.~\ref{fig:unmasked_voronoi}. Note, that this Voronoi representation has no information of the land region where there should be no values. This problem is addressed in next section.

% \begin{figure}
%     \centering
%     \includegraphics[width=1\linewidth]{images/Voronoi_Example_Image.png}
%     \caption{Work flow of Voronoi Tessellation. The ground truth dataset contains sensor locations (a), these sensor read a value from our ground truth (b), the Voronoi Tessellation is used to make a dense representation (c)}
%     \label{fig:3.2}
% \end{figure}

\subsection{Land Mask}

Section 3.1 detailed the general Voronoi, and this case works for most problems, but it lacks information about the land region if there exists one. Thus in domain specific areas we potentially lose valuable information like where land regions are. 

% by incorporating  domain knowledge for reconstructing sea surface temperatures 

\begin{figure}[t]
\begin{subfigure}{\linewidth}
    \centering
    \includegraphics[width = 0.594\linewidth]{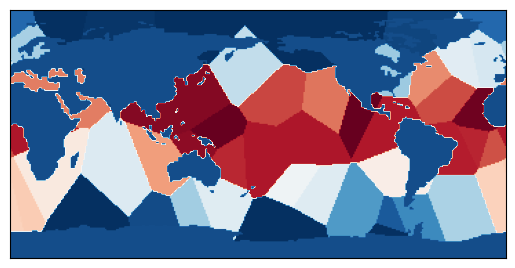}
    \caption{Masked Voronoi Representation \cite{Fukami2021}}
    \label{fig:masked_voronoi}
\end{subfigure}
\begin{subfigure}{.5\linewidth}
    \centering
    \includegraphics[width = \linewidth]{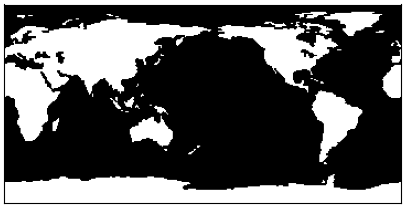}
    \caption{Land Mask}
    \label{fig:land_mask}
\end{subfigure}% 
\begin{subfigure}{.5\linewidth}
    \centering
    \includegraphics[width = \linewidth]{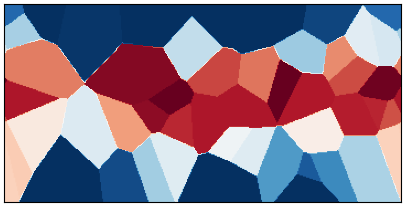}
    \caption{Unmasked Voronoi Representation}
    \label{fig:unmasked_voronoi}
\end{subfigure}
\caption{Separating the (a) Masked Voronoi into a (b) Land Mask and an (c) Unmasked Voronoi}
\label{fig:3.3}
\end{figure}

The paper we extend \cite{Fukami2021} solved this problem by overlaying a land mask on the Voronoi image, depicted in Fig.~\ref{fig:masked_voronoi}. Values were set to zero to signify continental land masses, but this approach creates ambiguous input as those values could be interpreted as actual sensor values. We argue that this ambiguity could reduce model performance. 

Thus, we propose to separate the masked Voronoi image input into two different images, the land mask and the Voronoi representation, to better inform the model. The separated land mask will be a binary mask, where 1 represents locations that need to be predicted and 0 represents locations that do not need to be predicted as seen in Fig.~\ref{fig:land_mask}.

\subsection{Distance Transform}

%%%%%%%%%%%%%%%%%%%%%%%%%%%%%%
%%% DT Imaged
%%%%%%%%%%%%%%%%%%%%%%%%%%%%%%
\begin{figure}[t]
\begin{subfigure}{.5\linewidth}
    \centering
    \includegraphics[width = \linewidth]{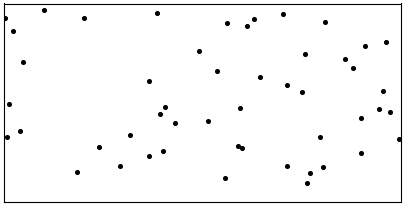}
    \caption{Sparse Location Mask \cite{Fukami2021}}
    \label{fig:sparse_loc_mask}
\end{subfigure}% 
\begin{subfigure}{.5\linewidth}
    \centering
    \includegraphics[width = \linewidth]{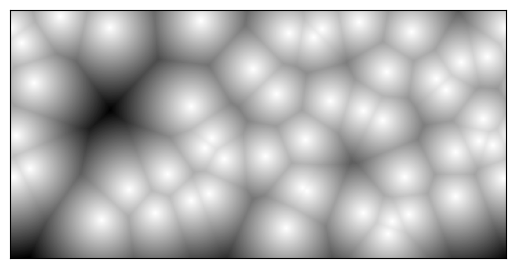}
    \caption{DT Information Mask}
    \label{fig:3.4.b}
\end{subfigure}

\begin{subfigure}{.5\linewidth}
    \centering
    \includegraphics[width = \linewidth]{images/noaa_land_binary_cropped.png}
    \caption{Land Mask}
    \label{fig:3.4.c}
\end{subfigure}
\begin{subfigure}{.5\linewidth}
    \centering
    \includegraphics[width = \linewidth]{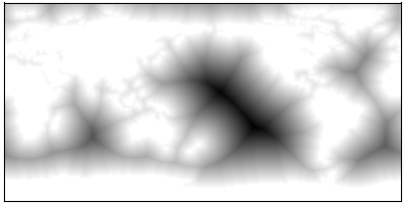}
    \caption{DT Land Mask}
    \label{fig:dt_land_mask}
\end{subfigure}
\caption{We apply the distance transform algorithm to sparse image representations such as the (a) Sparse Location Mask and the (c) Land Mask to achieve the dense (b) DT Information Mask and (d) DT Land Mask}
\label{fig:3.4}
\end{figure}

Using just the Land Mask and Voronoi representation is not sufficient for accurate spatial field reconstruction. As stated in \cite{Fukami2021}, there also needs to be information regarding the precise position of each sensor, a sensor location mask. In \cite{Fukami2021}, their sensor location mask, depicted in Fig.~\ref{fig:sparse_loc_mask}, contained only the exact locations of sensors and provided no information about how close the nearest sensor is from any given point in the spatial extents. Given that this image is constructed of mostly 0 values, the initial few convolution layers in a CNN will mostly see windows of all 0. Thus, this very sparse image is likely sub-optimal and could be improved.

To incorporate more information into the sparse location mask, we post-processed this image using the Distance Transform (DT). Given an image containing only 0 and 1 values, this operation computes the distance to the nearest 1 pixel for every 0 pixel in the image, where the distance can be $L_1$, $L_2$, and more. For our distance metric, we used the $L_2$ distance for the NOAA, Cylinder, and Channel datasets, described in detail  later, and used the haversine distance for the Antarctic dataset in the application portion of this paper. By using the DT, we now enable much more information about both the exact location of a sensor and the distance to the nearest sensor (if not an exact location). An example DT information mask is shown in Fig.~\ref{fig:3.4.b}. The following experiments will compare the use of a DT information mask or a sparse location mask, but never use both simultaneously.

This spatial information is not limited to sensors though, as a similar method can be applied to other inputs like the Land Mask. Denoting the distance a specific point is from land could potentially be useful information for a model. An example of the DT Land Mask can be visualized in Fig.~\ref{fig:dt_land_mask}. Experiments with the DT Land Mask were completed for the NOAA dataset but they did not improve reconstruction results and thus were not included in this work. While fruitless for the NOAA dataset, other reconstructions could benefit from from a DT Land mask if there is a spatial dependence between points and the land region.

\subsection{Circular Images for Circular Datasets}
\label{circ_section}

\begin{figure}[ht]
    \centering
    \includegraphics[width=1\linewidth]{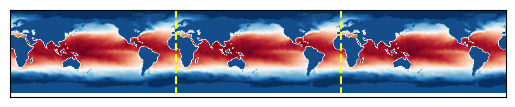}
    \caption{Circular Image Dataset Calculation}
    \label{fig:3.5}
\end{figure}

\begin{figure}[t]
\begin{subfigure}{.5\linewidth}
    \centering
    \includegraphics[width=\linewidth]{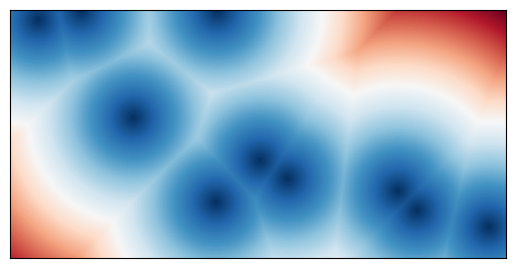}
    \caption{DT No Circular Method}
    \label{fig:3.6.a}
\end{subfigure}
\begin{subfigure}{.5\linewidth}
    \centering
    \includegraphics[width=\linewidth]{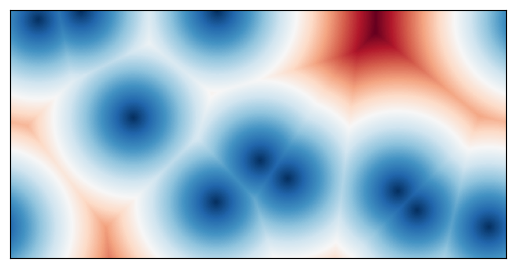}
    \caption{DT With Circular Method}
    \label{fig:3.6.b}
\end{subfigure}% 
\caption{Circular DT Correction}
\label{fig:3.6}
\end{figure}

The Distance Transform and Voronoi algorithms inherently work on a 2-D plane, but their methods could also be extended to 3-D representations as well. An example can be seen from the NOAA Dataset as it describes the sea surface temperature of the Earth's oceans, which is a 3-D dataset compressed into a 2-D representation. The original data was spherical, but when compressed to a 2-D representation, the information is distorted. Most notably the left and right hand edges of the 2-D representation seem disjoint, but in actuality they are adjacent on the 3-D sphere. The circular nature of the dataset is not taken into account by the naive Distance Transform and Voronoi algorithms. To solve this we implemented a Circular Image Dataset which runs the algorithms on an expanded image as seen in Fig.~\ref{fig:3.5}. By replicating the image two more times on both sides, we allow the image to be correctly computed and connected in a circular and continuous manner. Note, the circular representation is a single solution to the problem of correctly creating the Distance Transform and Voronoi for a 3-D dataset and a second more robust solution is described in the applications section for the Antarctic reconstruction. The circular image method, while inferior to the method described later, was described because all NOAA models where trained using this approach.
A visualization of the improvement using a circular method can be seen in  Fig.~\ref{fig:3.6}. The improvements are obvious in the upper right and lower left corners of the images. 


\subsection{Masked Out Loss}

The final domain specific improvement that can be made is masking out the loss of the land region during model training. The land region does not need to be accurately reconstructed since it is masked out in post-processing, thus including the loss of reconstructing the land is not useful and can lead to greater computation time.

\subsection{Normalization}

\label{section:Normalization}

Lastly, we propose using normalization for all input images and ground truth images for a model, which \cite{Fukami2021} did not do. We normalized all pixels in an image to be in the range [-1,1] by dividing each pixel value by the absolute maximum pixel value across all training images. Normalization was proposed to ensure all input and ground truth images were in the same small value range as normalization can improve model training stability, training speed, and overall model performance \cite{lecun-98, lecun-98b}. With normalization, predicted reconstructions will be roughly confined but not limited to the range [-1,1]. True predictions for real values can be recovered by multiplying the prediction by the constant used during normalization.

% Dataset Section
\section{Datasets}
\label{datasets.ch}

\begin{figure}[t]
\centering
\begin{subfigure}{.33\linewidth}
    \includegraphics[width = 0.9\linewidth]{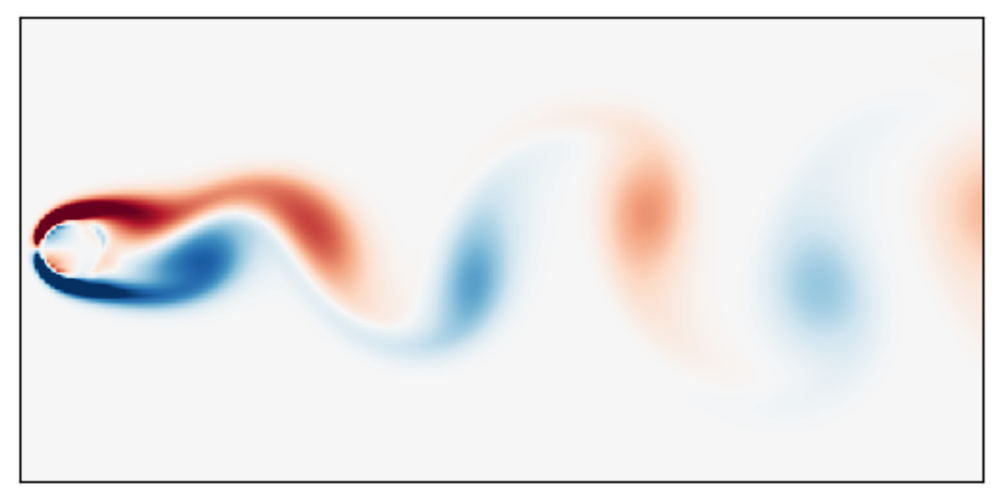}
    \caption{}
    \label{fig:cylinder_gt}
\end{subfigure}% 
\begin{subfigure}{.33\linewidth}
    \includegraphics[width = 0.9\linewidth]{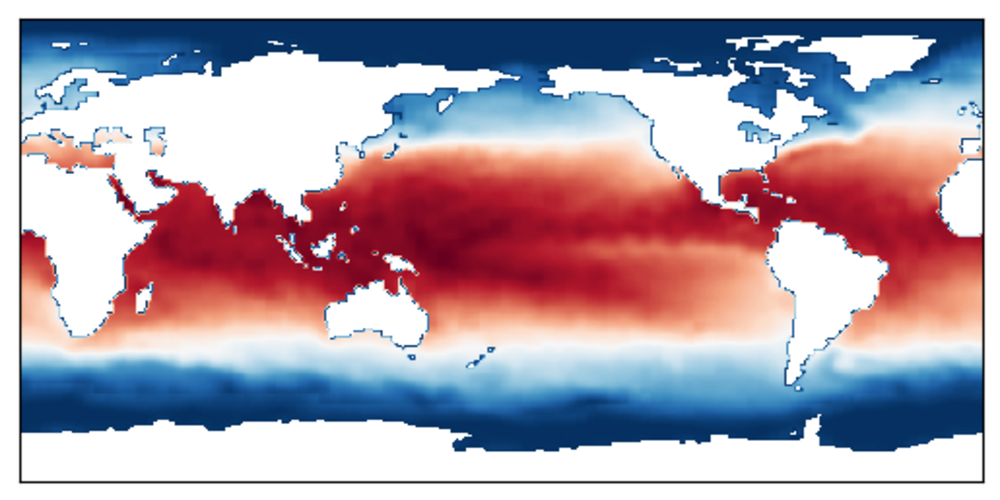}
    \caption{}
    \label{fig:noaa_gt}
\end{subfigure}% 
\begin{subfigure}{.33\textwidth}
    \includegraphics[width = 0.9\linewidth]{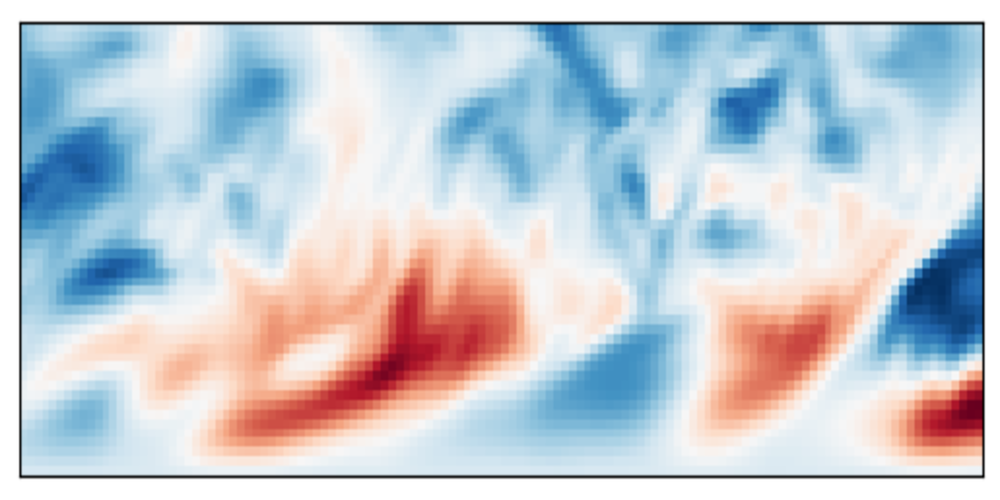}
    \caption{}
    \label{fig:channel_gt}
\end{subfigure}
\caption{Example images from the a) Cylinder, b) NOAA, and c) Channel datasets}
\label{fig:gt_figs}
\end{figure}

% Include images of the datasets

We evaluated our approach on the same three datasets examined in \cite{Fukami2021}: two-dimensional cylindrical wake \cite{Taira2007, Colonius2007}, NOAA sea surface temperature \cite{Reynolds2002}, and turbulent channel flow \cite{Fukagata2006}. Examples from these datasets are shown in Fig.~\ref{fig:gt_figs}. Descriptions of the final dataset we apply our method to will be described in the applications section. 

\subsection{Cylinder}

The cylinder wake dataset (Cylinder) is a common canonical example of fluid flow that is used throughout fluid dynamics. It consists of 5,000 single-channel images each with a resolution of 94x189 pixels representing the rotation speed of a fluid flowing around a cylinder. The cylinder dataset is a cyclical dataset as it repeats vortex shedding roughly every 1,200 images. Thus, the cylinder dataset covers roughly 4 complete cycles. We show that even though there are few cycles in a dataset, our methodology still improves  reconstruction performance.

\subsection{NOAA}

The NOAA sea surface temperature dataset (NOAA) consists of 1,914 single-channel images each with a resolution of 180x360 pixels representing sea surface temperature in Celsius. The NOAA dataset images are weekly observations of the sea surface temperature with a spatial resolution of $1^{\circ}$ latitude by $1^{\circ}$ longitude. The NOAA dataset is also cyclical in nature due to the pattern of seasonal temperature changes. Thus, there exists seasonal cycles every year, which is every 52 images. It should be noted that the NOAA dataset is unique in the sense that it has a land region where no measurements are taken. Through the NOAA dataset we show how to handle datasets that have masked out regions devoid of observations and show our approach can extend beyond just spatial sensor locations.

\subsection{Channel}

The turbulent channel flow dataset (Channel) consists of 10,000 single-channel 48x128 pixel images where each pixel represents fluid velocity. Unlike the NOAA and Cylinder datasets, the Channel dataset is chaotic and has no reoccurring cycle within the data. This last dataset exhibits the challenges that deep learning has for reconstructing chaotic datasets in general and details a case where using the distance transform and knowing the spatial relation of sensors is not useful.

\subsection{Data Splits}

For image reconstruction, there are two types of prediction paradigms: interpolation and extrapolation. In interpolation, there are gaps temporally within the training set making the goal of reconstruction to fill the gaps between the training data. In extrapolation, training is done on a complete time series of images to perform image reconstruction at future time points. Both paradigms are important, but for the sake of simplicity we only focus on interpolation to be consistent with past papers in reconstruction \cite{Erichson2020, Fukami2021}. Since we are testing interpolation, our train/val/test split must create gaps temporally within the data.

% I want to include some part about how the information in our model is not dependent on time aswell.
While papers such as \cite{Erichson2020, Fukami2021}, used random sampling for splitting their data, we completed a partitioned random split of the data. We argue that pure random sampling could leak training information into the test and validation sets since temporally adjacent images are very similar to each other. This could cause evaluation metrics to overstate good performance during training and validation, but result in significantly degraded performance during deployment. Thus, we randomly sampled chunks of data rather than single data examples themselves, hence a partitioned random split. The motivation of this methodology is to limit the amount of times a sample in the training set is temporally adjacent to samples in the validation set, which pure random sampling does not achieve. Furthermore, by sampling chunks rather than sampling single images, we are more likely to capture the cyclical nature of a dataset in both training and testing.

% Dataset Results
\section{Results and Experiments}
\label{results.ch}

The following section details the evaluation of our proposed methodology. We start by describing our model's training and evaluation setup followed by experiments with the three aforementioned datasets. Each experiment with a dataset will be comprised of three parts: 1) A table of test errors, 2) a graph of validation error, and 3) several example reconstructions.

\subsection{Model Training}
\label{sect:model_training}

Training inputs are constructed by choosing $n$ random points as sensor locations. The sparse location mask, DT information mask, and Voronoi representation are subsequently derived from these points. For every distinct amount of $n$ sensors, we construct multiple arrangements of these $n$ sensors to create unique sets of sensor placements. By training on unique sets of sensor locations, both in placement and in sensor quantity, a single machine learning model is able to be robust to sensors moving and going online / offline. Thus to ensure generalization in our models to different sensor configurations, we trained on a set of different sensor amounts and sensor placements for each dataset. The NOAA training dataset is comprised of 25 unique cases of sensor amounts and placements. The number of sensors used comes from the set $n_{sensor, train, noaa} = \{10,20,30,50,100\}$ with the random seeds used to generate sensor placements coming from the set $n_{seed, train} = \{1,2,100,200,300\}$. The Cartesian product $n_{sensor, train}$ x $n_{seed, train}$ denotes the list of all possible cases on which the NOAA reconstruction model is trained. Similarly the Cylinder dataset uses sensor amounts from the set $n_{sensor, train, cylinder} = \{10,25,50,75,100\}$ and the Channel dataset uses sensor amounts from the set $n_{sensor, train, channel} = \{50,100,200\}$. Both the Cylinder and Channel datasets use random seeds from the same $n_{seed, train}$ set as the NOAA dataset.

All experiments were trained using a simple 8-layer CNN following the same architecture as in \cite{Fukami2021}. All models were trained using ADAM optimization for a maximum of 300 epochs using a learning rate of $1e^{-4}$ with a batch size of 32. The epoch with the best validation accuracy was used to select the final model. We used MSE loss for training error and reported relative $L_2$ error as the metric for reconstruction performance, which is consistent with \cite{Fukami2021}. Relative $L_2$ error is computed from $\epsilon = ||Y - \hat{Y}||_2/ ||Y||_2$ where $Y$ is the ground truth reconstruction and $\hat{Y}$ is the predicted reconstruction from the model. For experiments involving normalization, images were normalized to be in the range [-1,1] as mentioned in Sect.~\ref{section:Normalization}. All models were coded using PyTorch and trained with the same initialization seed of 1 for a fair assessment.

\subsection{Model Evaluation}

To evaluate the model for cases when sensor locations are fixed and other scenarios when sensor locations may have changed from their initial placement, we create two types of tests. The first is the ``seen'' sensor test, which tests the trained model on examples where the sensor amounts and locations match that of the training set. The second is the ``unseen'' sensor test, which tests the trained model on examples where the sensor placements are completely different from those during training. This involved both different seeds for sampling sensor placements and different sensor amounts for which the model was not trained on. Good performance on the unseen sensor amount dataset would show that a model is robust to error even when sensors are moving or coming online/offline overtime.

The following tables displaying experimental results provide the relative $L_2$ reconstruction error for varying amounts of sensors and for both seen and unseen sensor placements. The ``Experiment'' column denotes the information of the experiment and tells us what modifications were made to the input data. The ``Seen Sensor Amounts'' and ``Unseen Sensor Amounts'' sections have columns denoting the amount of sensors that were tested.  Each reported score for a given experiment and sensor amount is the average relative $L_2$ error across multiple sensor placements. For the seen case, we report the average relative $L_2$ error for the 5 sensor placements used during training, and in the unseen case we report the average relative $L_2$ error for 5 sensor placements not used during training. We tested multiple sensor placements (seeds) in our results to ensure they were not biased towards a single sensor placement.

% For experimentation, all models were trained on datasets that contained multiple different sensor placements with varying amount of sensors to ensure generalizability. To accomplish this, we utilized 5 different random seeds for sampling the locations of sensors ($seeds = {300,200,100,1,2}$). Each data set had its own subset of sensor amounts since each dataset is different in size. For the Cylinder dataset, the number of sensors for training is set to $n_{sensor,train} = {10,20,30,50,100}$. For the NOAA dataset, the number of sensors for training was set to $n_{sensor,train} = {10,25,50,75,100}$. For the Channel dataset, the number of sensors for training was set to $n_{sensor,train} = {50,100,200}$. It should be noted that a single machine learning model is trained and used for all combinations of sensor amounts and sensor placements.

\newpage

%%%%%%%%%%%%%%%%%%%%
%%%%% Cylinder %%%%%
%%%%%%%%%%%%%%%%%%%%
\subsection{Cylinder Dataset Results}

\begin{table}[b]
\caption{Cylinder experiment results}
\scriptsize
\begin{center}
\begin{tabular}{l |c c c |c c c c c}
\hline
Experiment & \multicolumn{3}{c|}{Seen Sensor Amounts} & \multicolumn{5}{c}{Unseen Sensor Amounts}\\
\hline
 & 10  & 50 & 100 & 10 & 35 & 50 & 85 & 100 \\
\hline
\hline
Baseline & 0.155 & 0.043 & 0.037 & 1.030 & 0.931 & \textbf{0.829} & 0.785 & 0.778 \\
\hline
Normalized & 0.171 & 0.043 & 0.036 & \textbf{1.016} & 0.922 & 0.838 & 0.923 & 0.810 \\
DT mask & 0.167 & 0.044 & 0.036 & 1.025 & 0.918 & 0.855 & 0.807 & 0.791 \\
Normalized \& DT mask & \textbf{0.119} & \textbf{0.035} & \textbf{0.030} & 1.033 & \textbf{0.908} & 0.847 & \textbf{0.780} & \textbf{0.774} \\
\hline
\end{tabular}
\end{center}
\end{table}

We first examine reconstruction on the Cylinder dataset. In Table 1, results of using normalization, a DT information mask, and the combination of both are compared to the baseline method of \cite{Fukami2021}, which simply uses a sparse location mask and a Voronoi representation. When using only normalization or the DT mask, we see insignificant differences from the baseline approach. However when both are combined, we see significant improvement when testing on the seen sensor locations compared to the original approach. While a normalized model using a DT mask is optimal for seen sensor placements, there seems to be no optimal model for the unseen case. This likely means it is crucial for the model to train and evaluate on a fixed set of sensor locations for the cylinder dataset as changing sensor placements are detrimental to the accuracy of the reconstruction no matter what method is used. The cylinder reconstruction experiments show that normalization and a DT information mask are beneficial for reconstruction performance when sensors are fixed.

\subsubsection{Cylinder Validation Curves}

\begin{figure}[ht]
    \centering
    \includegraphics[width = 0.7\linewidth]{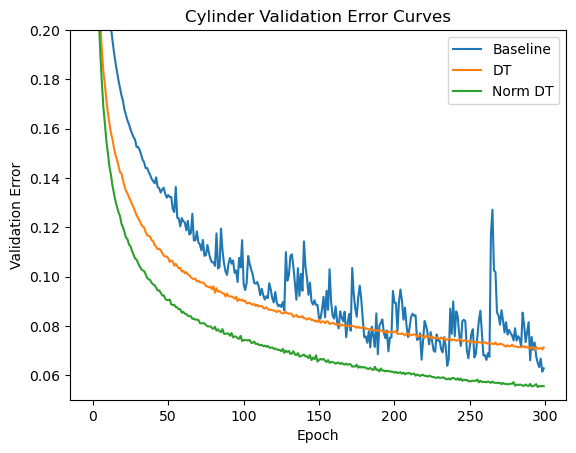}
\caption{Cylinder validation error plots comparing the baseline approach to two models using our methods}
\label{cyl_val_curve}
\end{figure}

While using the DT mask and normalization improves accuracy, the more important benefit of these data augmentations is they allow for faster and stabler training. As seen in the Cylinder validation error curves depicted in Fig.~\ref{cyl_val_curve}, the baseline validation error is noisy while the use of DT information masks promotes a stable learning curve. Furthermore, a model using a DT information mask with normalization achieves the optimal training performance of the baseline model in 100 less epochs, denoting that our augmentations improve training speed significantly. 

\subsubsection{Cylinder Reconstruction Examples}

The following pages provide examples of spatial field reconstructions for the Cylinder dataset. Each reconstruction diagram for the following images, and for all dataset reconstructions, contain three parts: 1) the inputs of the model in the first row, 2) the predicted reconstruction next to the ground truth image in the second row, and 3) two images depicting the error of the reconstruction in the third and final row. The first error map depicts the squared difference between the ground truth and image reconstruction model and is unbounded to show the maximal error of a reconstruction. The second error map, termed scaled error, is a capped error map with a power-law scaled colorbar used to represent the general patterns of error that occur throughout the reconstruction image at low levels of error. The scaled error map is useful to compare reconstructions between models. Note that all images in these examples contain black or white dots that represent the location of the sensors. These black/white dots are not part of the true image and are merely for reference.

\begin{figure}[!ht]
% Inputs
\begin{subfigure}{.49 \linewidth} 
    \centering
    \includegraphics[width = \linewidth]{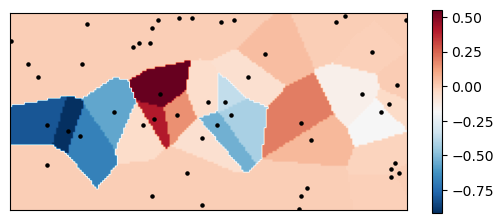}
    \caption{Voronoi}
\end{subfigure}
\begin{subfigure}{.49 \linewidth}
    \centering
    \includegraphics[width = \linewidth]{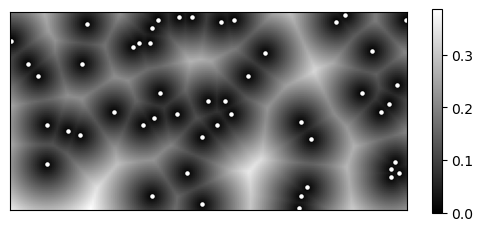}
    \caption{DT Mask}
\end{subfigure}% 

\vspace{40pt}

% Prediction
\begin{subfigure}{0.49\linewidth}
    \centering
    \includegraphics[width = \linewidth]{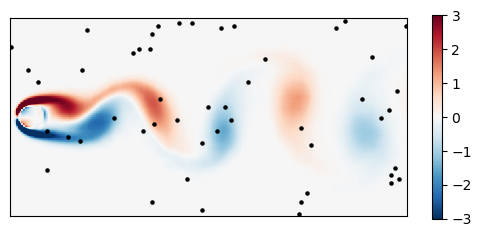}
    \caption{Image Reconstruction (Prediction)}
\end{subfigure} %
\begin{subfigure}{0.49 \linewidth}
    \centering
    \includegraphics[width = \linewidth]{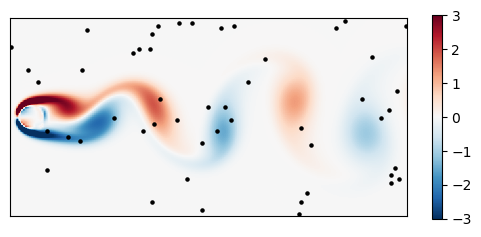}
    \caption{Ground Truth}
\end{subfigure}

\vspace{40pt}

% Forth Row of images
\begin{subfigure}{0.49 \linewidth}
    \centering
    \includegraphics[width = \linewidth]{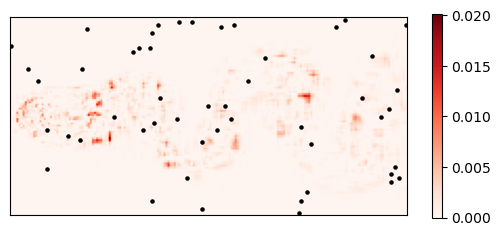}
    \caption{Full Error Map}
\end{subfigure}
\begin{subfigure}{0.49 \linewidth}
    \centering
    \includegraphics[width = \linewidth]{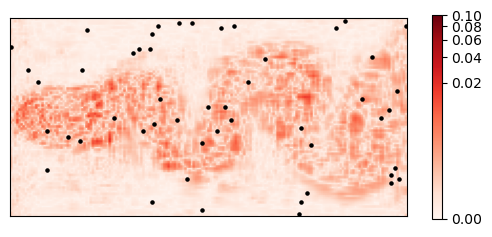}
    \caption{Scaled Error Map}
\end{subfigure}

\caption{Cylinder dataset reconstruction example for 50 seen sensors using normalization and a DT mask in the model}
\label{fig:cylinder_dt_full_seen}
\end{figure}

\begin{figure}[!ht]
% Inputs
\begin{subfigure}{.49 \linewidth} 
    \centering
    \includegraphics[width = \linewidth]{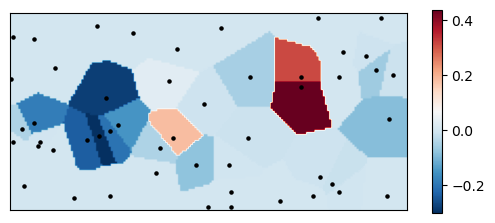}
    \caption{Voronoi}
\end{subfigure}
\begin{subfigure}{.49 \linewidth}
    \centering
    \includegraphics[width = \linewidth]{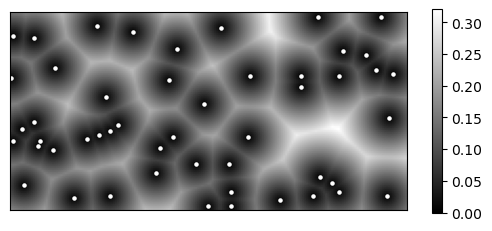}
    \caption{DT Mask}
\end{subfigure}% 

\vspace{40pt}

% Prediction
\begin{subfigure}{0.49\linewidth}
    \centering
    \includegraphics[width = \linewidth]{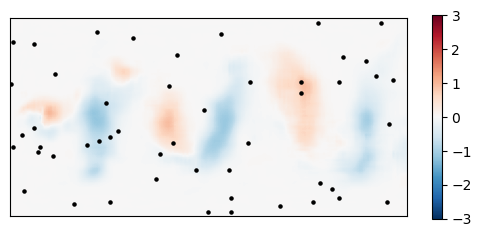}
    \caption{Image Reconstruction (Prediction)}
\end{subfigure}%
\begin{subfigure}{0.49\linewidth}
    \centering
    \includegraphics[width = \linewidth]{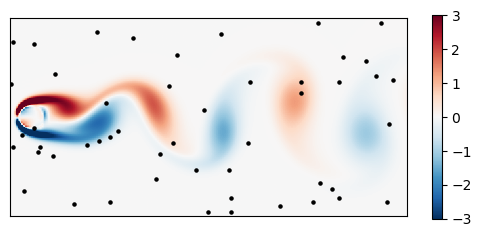}
    \caption{Ground Truth}
\end{subfigure}

\vspace{40pt}

% Forth Row of images
\begin{subfigure}{0.49 \linewidth}
    \centering
    \includegraphics[width = \linewidth]{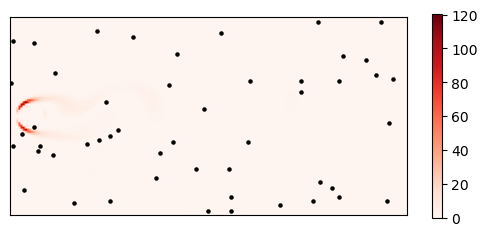}
    \caption{Full Error Map}
\end{subfigure}
\begin{subfigure}{0.49 \linewidth}
    \centering
    \includegraphics[width = \linewidth]{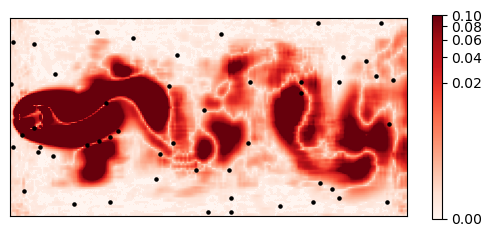}
    \caption{Scaled Error Map}
\end{subfigure}

\caption{Cylinder dataset reconstruction example for 50 unseen sensors using normalization and a DT mask in the model}
\label{fig:cylinder_dt_full_unseen}
\end{figure}

The first two reconstruction examples for the Cylinder dataset compare reconstructions under seen and unseen sensor locations using the normalized model with a DT mask, which was the optimal model for the seen sensor experiments. The reconstructions under the seen sensor experiments represented in Fig.~\ref{fig:cylinder_dt_full_seen} are extremely accurate with the error maps showing negligible error and the predicted image reconstruction appearing identical to the ground truth image. Under the unseen sensor experiment (Fig.~\ref{fig:cylinder_dt_full_unseen}), reconstruction was not successful and does not resemble the ground truth. The reconstruction under the unseen case has no generalization capability and produces cylinder vortex flow fields that appear to be Gaussian blurred versions of the Voronoi Input. 

These visual examples demonstrate the challenges of reconstructing unseen sensor placements for the Cylinder dataset. The hypothesis for our model's shortcomings is due to the lack of training data for a given cycle. Since this dataset has only 4 cycles, two of which are for testing and validation, there is little data for the model to understand generalization when sensors move. Thus with limited temporal data, the model might overfit to perform well on a seen set of sensors and under-perform when these sensors move.

\clearpage

%%%%%%%%%%%%%%%%%%%%%
%%% Cylinder Images
%%%%%%%%%%%%%%%%%%%%%

%%%% Second Example
\newpage
\clearpage

\newpage

%%%%%%%%%%%%%%%%%%%%
%%%%% Noaa %%%%%
%%%%%%%%%%%%%%%%%%%%
\subsection{NOAA Dataset Results}

\begin{table}[b]
\caption{NOAA seen experiment results}
\scriptsize
\begin{center}
\begin{tabular}{l |c c c | c c c c c}
\hline
Experiment & \multicolumn{3}{c}{Seen Sensor Amounts} & \multicolumn{5}{c}{Unseen Sensor Amounts}\\
\hline
 & 10 & 50 & 100 & 10 & 50 & 70 & 100 & 200 \\
\hline
\hline
Baseline & 0.071 & 0.046 & 0.041 & 0.311 & 0.139 & 0.120 & 0.101 & 0.087\\
\hline
Normalized & 0.070 & 0.045 & 0.040 & 0.313 & 0.134 & 0.118 & 0.098 & 0.081\\
DT mask & 0.066 & 0.046 & 0.041 &  0.299 & 0.107 & 0.100 & 0.088 & 0.072\\
Separate Voronoi & 0.075 & 0.048 & 0.044 & \textbf{0.247} & 0.127 & 0.110 & 0.098 & 0.079\\
Normalized \& DT mask & 0.064 & 0.045 & 0.041 & 0.302 & 0.110 & 0.097 & 0.083 & 0.067\\
Separate Voronoi \& Masked loss & 0.068 & 0.045 & 0.040 & 0.297 & 0.132 & 0.114 & 0.097 & 0.081\\
Normalized \& DT mask \& Separate Voronoi & 0.064 & 0.044 & 0.039 & 0.267 & 0.095 & 0.082 & \textbf{0.073} & \textbf{0.056} \\
Normalized \& DT mask \& Separate Voronoi \& Masked loss & \textbf{0.062} & \textbf{0.043} & \textbf{0.038} & 0.258 & \textbf{0.093} & \textbf{0.081} & \textbf{0.073} & 0.057\\
\hline
\end{tabular}
\end{center}
\label{table_noaa}
\end{table}

% The best scores are emphasized in bold.

Next we examine reconstruction on the NOAA dataset. In Table~\ref{table_noaa}, we compare the results of using DT information masks, data normalization, a loss function with the land region masked out, separating the land mask from the masked Voronoi, and varying combinations of these methods against the baseline approach. We do not experiment with only using the masked loss as we view the masked loss as a supplemental change associated with the separation of the Voronoi. In Table~\ref{table_noaa}, we show that a combination of all four proposed augmentations is the optimal model.

We found that separating the land mask from the masked Voronoi and normalization alone did not improve performance compared to the baseline in the seen case, but separating the masked Voronoi did improve the performance in all unseen sensor placement experiments. Separating the masked Voronoi alone did achieve the optimal performance across all experiments when using 10 unseen sensor amounts, but this success is attributed to the variability that comes with significantly sparse data. Notably, using the DT mask alone was able to improve or maintain model performance in all seen and unseen cases. If a single augmentation were to be used over others, it would be the DT mask. To stay consistent with Cylinder, we experimented using both Normalization and the DT mask and found it outperformed the baseline as expected, but was not significantly different from just using the DT mask alone. Next, we found that when we normalized our data, used a DT information mask, and separated the masked Voronoi, significant increases in reconstruction accuracy were observed across all cases, most notably the unseen test cases. When masked loss was incorporated to the previously mentioned experiment, there was a minor increase in reconstruction accuracy. The NOAA reconstruction experiments show that using normalization, DT masks, and careful application of domain knowledge can all improve reconstruction performance.

\subsubsection{NOAA Validation Curves}

\begin{figure}[ht]
    \centering
    \includegraphics[width = 0.8\linewidth]{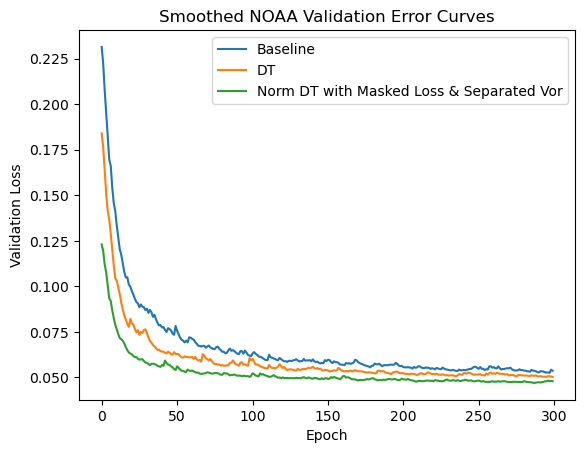}
\caption{NOAA validation error plots comparing the baseline approach to two models using our methods}
\label{noaa_val_curve}
\end{figure}

% Need to complete wording on this one
Similar to Cylinder, our methods improved training speed for NOAA reconstructions as seen in Fig.~\ref{noaa_val_curve}. While many combinations of data augmentations were tested, only the validation error curves for the baseline, DT, and optimal model were shown for simplicity. Furthermore, validation errors were smoothed using an exponential moving average to focus on showing the overall trend in training speed. 

% Finish correcting this section. Needs refinement
The validation curves in Fig.~\ref{noaa_val_curve} depict that the optimal model, which used a separated Voronoi, masked loss, DT mask, and normalization, achieves a lower validation loss after 50 epochs than the baseline model after 300 epochs. This is a six times speed up in training speed to reach the same performance. Furthermore, solely using a DT mask instead of a Sparse Location Mask improves training speed by roughly 75 epochs compared to the baseline. These increases in computational speed are significant under our training conditions as a single epoch takes roughly 10 minutes of training. Thus, a six times speed up in computation reduces training by days.

\subsubsection{NOAA Reconstruction Examples}

\begin{figure}[!htb]
% Inputs
\begin{subfigure}{.33\linewidth}
    \centering
    \includegraphics[width = \linewidth]{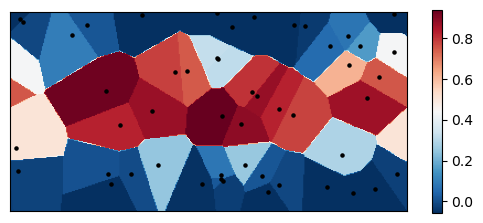}
    \caption{Voronoi}
\end{subfigure}
\begin{subfigure}{.33\linewidth}
    \centering
    \includegraphics[width = \linewidth]{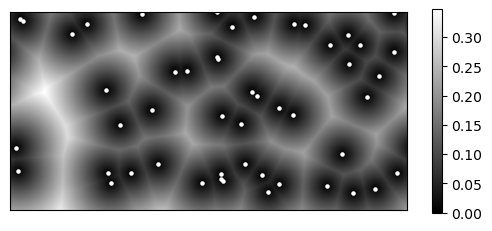}
    \caption{DT Mask}
\end{subfigure}% 
\begin{subfigure}{.33\linewidth}
    \centering
    \includegraphics[width = \linewidth]{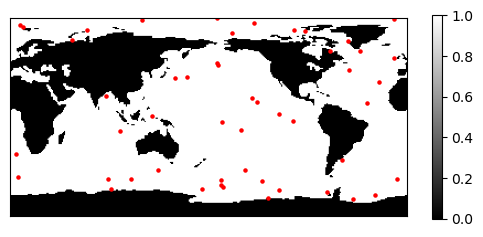}
    \caption{Land Mask}
\end{subfigure}

\vspace{40pt}

% Prediction
\begin{subfigure}{0.49\linewidth}
    \centering
    \includegraphics[width = \linewidth]{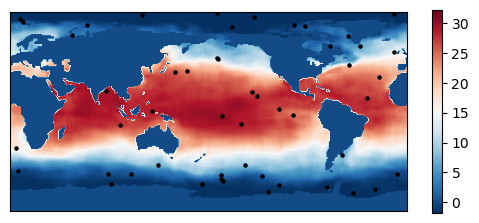}
    \caption{Image Reconstruction (Prediction)}
\end{subfigure} %
\begin{subfigure}{0.49\linewidth}
    \centering
    \includegraphics[width = \linewidth]{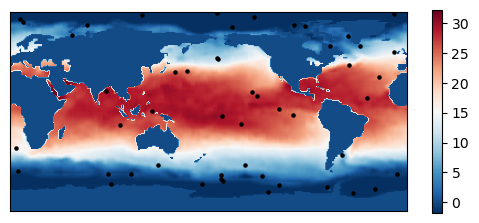}
    \caption{Ground Truth}
\end{subfigure}

\vspace{40pt}

% Forth Row of images
\begin{subfigure}{0.49 \linewidth}
    \centering
    \includegraphics[width = \linewidth]{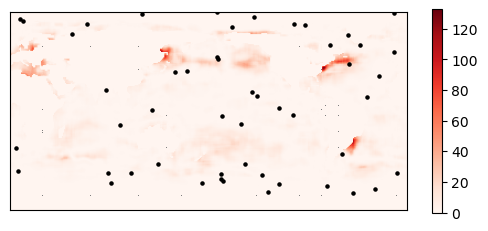}
    \caption{Error Map}
\end{subfigure}% 
\begin{subfigure}{.49 \linewidth}
    \centering
    \includegraphics[width = \linewidth]{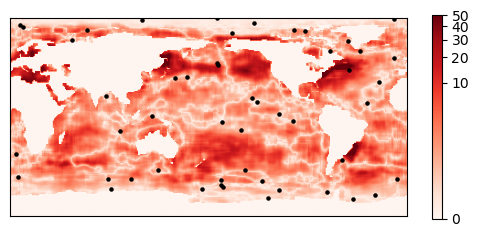}
    \caption{Scaled Error Mask}
\end{subfigure}

\caption{ NOAA dataset reconstruction example for 50 unseen sensors using optimal model, which uses normalization, DT Mask, Masked Loss, and a separated Voronoi}
\label{fig:unseen_50_optimal_recon}
\end{figure}

\begin{figure}[htbp]
% Inputs
\begin{subfigure}{.49\linewidth}
    \centering
    \includegraphics[width = \linewidth]{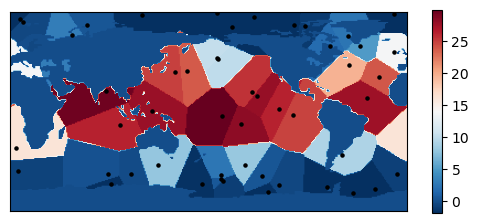}
    \caption{Voronoi}
\end{subfigure}
\begin{subfigure}{.49\linewidth}
    \centering
    \includegraphics[width = \linewidth]{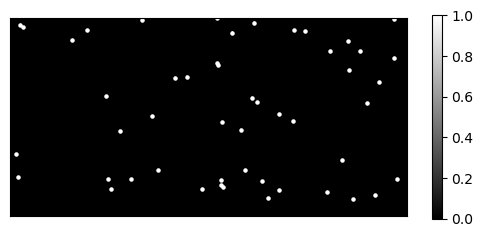}
    \caption{Sparse Mask}
\end{subfigure}

\vspace{40pt}

% Prediction
\begin{subfigure}{0.49 \linewidth}
    \centering
    \includegraphics[width = \linewidth]{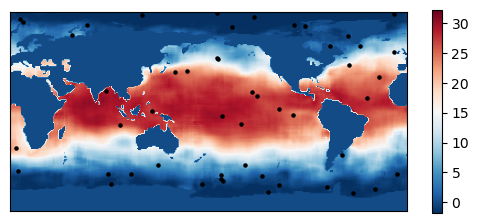}
    \caption{Image Reconstruction (Prediction)}
\end{subfigure} %
\begin{subfigure}{0.49\linewidth}
    \centering
    \includegraphics[width = \linewidth]{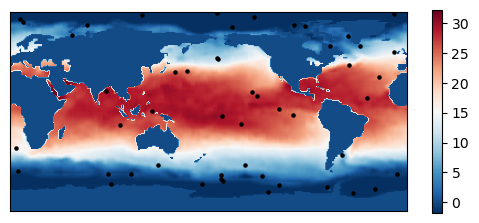}
    \caption{Ground Truth}
\end{subfigure}

\vspace{40pt}

% Forth Row of images
\begin{subfigure}{0.49 \linewidth}
    \centering
    \includegraphics[width = \linewidth]{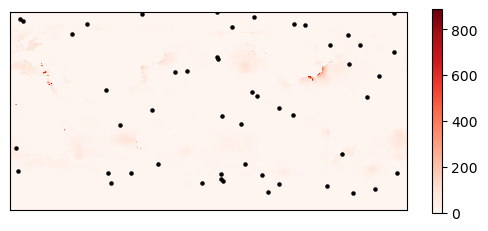}
    \caption{Error Map}
\end{subfigure}% 
\begin{subfigure}{.49 \linewidth}
    \centering
    \includegraphics[width = \linewidth]{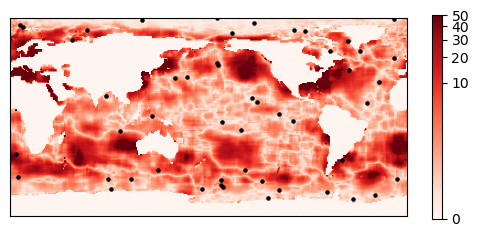}
    \caption{Scaled Error Mask}
\end{subfigure}

\caption{NOAA dataset reconstruction example for 50 unseen sensors using baseline model}
\label{fig:unseen_baseline_50_noaa}
\end{figure}

The first two reconstruction examples for the NOAA dataset in Fig.~\ref{fig:unseen_50_optimal_recon} and Fig.~\ref{fig:unseen_baseline_50_noaa} compare reconstructions from the optimal and baseline models using 50 unseen sensors. Comparing the prediction images of the models shows that the optimal model had smoother reconstructions than the baseline model which tended to have sporadic or noisy reconstructions in various locations. The error map further reveals that the max error for the baseline reconstruction is much greater than the max error for the optimal reconstruction. Overall, the optimal model outperforms the baseline with its smoother and more consistent reconstructions.

\begin{figure}[htbp]
% Inputs
\begin{subfigure}{.33\linewidth}
    \centering
    \includegraphics[width = \linewidth]{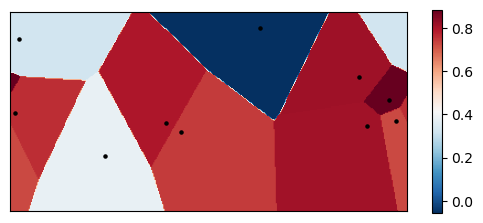}
    \caption{Voronoi}
\end{subfigure}
\begin{subfigure}{.33\linewidth}
    \centering
    \includegraphics[width = \linewidth]{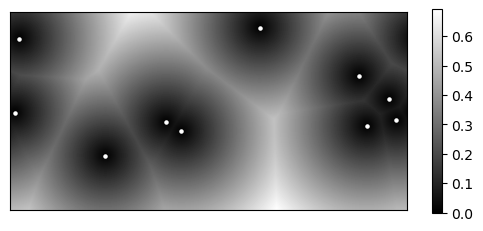}
    \caption{DT Mask}
\end{subfigure}% 
\begin{subfigure}{.33\linewidth}
    \centering
    \includegraphics[width = \linewidth]{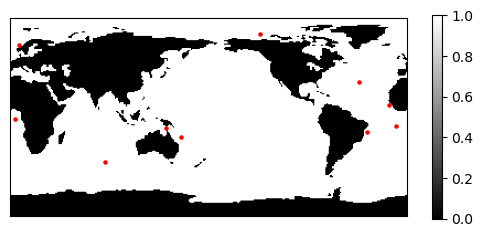}
    \caption{Land Mask}
\end{subfigure}

\vspace{40pt}

% Prediction
\begin{subfigure}{0.49\linewidth}
    \centering
    \includegraphics[width = \linewidth]{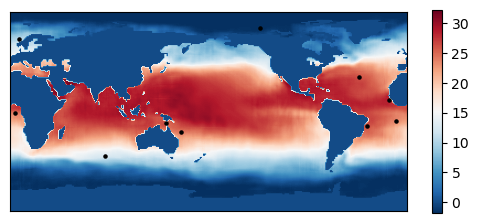}
    \caption{Image Reconstruction (Prediction)}
\end{subfigure} %
\begin{subfigure}{0.49\linewidth}
    \centering
    \includegraphics[width = \linewidth]{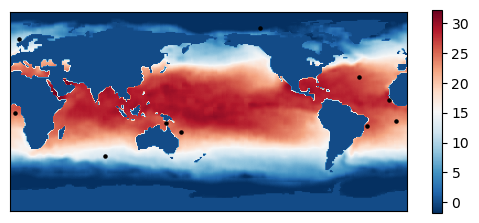}
    \caption{Ground Truth}
\end{subfigure}

\vspace{40pt}

% Forth Row of images
\begin{subfigure}{0.49 \linewidth}
    \centering
    \includegraphics[width = \linewidth]{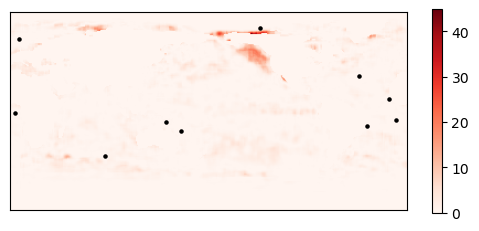}
    \caption{Error Map}
\end{subfigure}% 
\begin{subfigure}{.49 \linewidth}
    \centering
    \includegraphics[width = \linewidth]{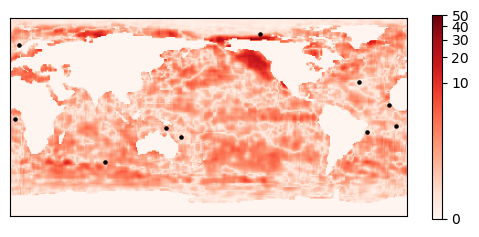}
    \caption{Scaled Error Mask}
\end{subfigure}

\caption{NOAA dataset reconstruction example for 10 seen sensors using optimal model}
\label{fig:seen_10_optimal_noaa}
\end{figure}

\begin{figure}[htbp]
% Inputs
\begin{subfigure}{.33\linewidth}
    \centering
    \includegraphics[width = \linewidth]{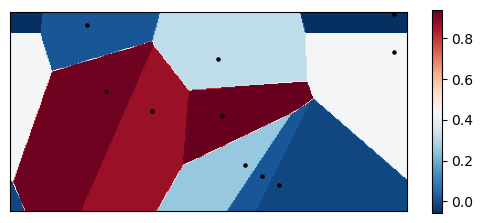}
    \caption{Voronoi}
\end{subfigure}
\begin{subfigure}{.33\linewidth}
    \centering
    \includegraphics[width = \linewidth]{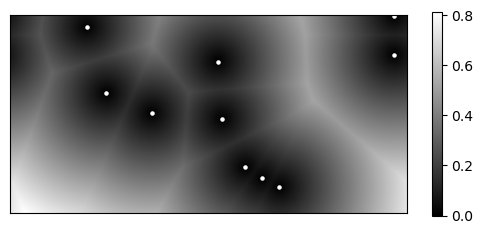}
    \caption{DT Mask}
\end{subfigure}% 
\begin{subfigure}{.33\linewidth}
    \centering
    \includegraphics[width = \linewidth]{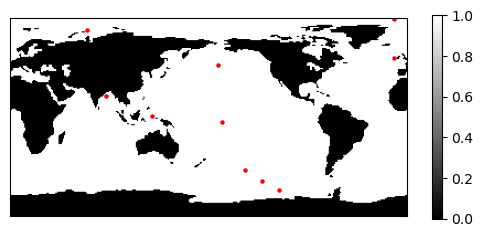}
    \caption{Land Mask}
\end{subfigure}

\vspace{40pt}

% Prediction
\begin{subfigure}{0.49\linewidth}
    \centering
    \includegraphics[width = \linewidth]{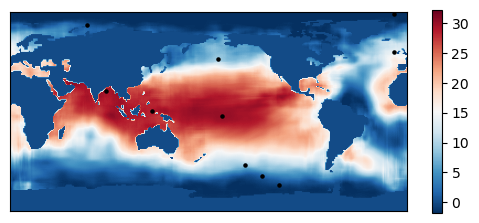}
    \caption{Image Reconstruction (Prediction)}
\end{subfigure} %
\begin{subfigure}{0.49\linewidth}
    \centering
    \includegraphics[width = \linewidth]{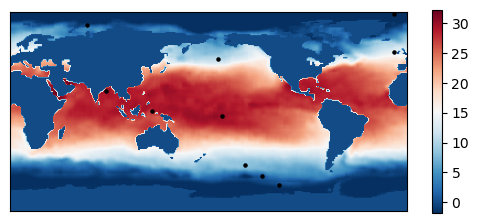}
    \caption{Ground Truth}
\end{subfigure}

\vspace{40pt}

% Forth Row of images
\begin{subfigure}{0.49 \linewidth}
    \centering
    \includegraphics[width = \linewidth]{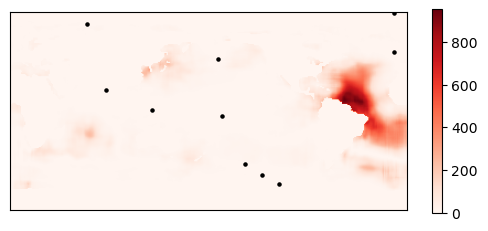}
    \caption{Error Map}
\end{subfigure}% 
\begin{subfigure}{.49 \linewidth}
    \centering
    \includegraphics[width = \linewidth]{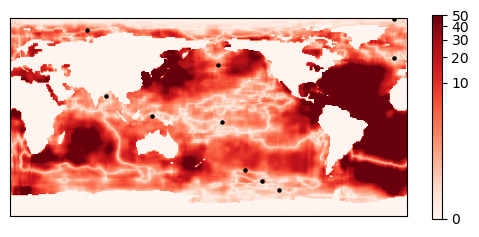}
    \caption{Scaled Error Mask}
\end{subfigure}

\caption{NOAA dataset reconstruction example for 10 unseen sensors using optimal model}
\label{fig:unseen_10_optimal_noaa}
\end{figure}

The last two examples for the NOAA dataset in Fig.~\ref{fig:seen_10_optimal_noaa} and Fig.~\ref{fig:unseen_10_optimal_noaa} compare reconstructions from 10 seen and 10 unseen sensors using the optimal model. Since the error rates are similar for the seen and unseen case for 50 sensors and more, using only 10 sensors showcases the difference between the seen and unseen reconstructions. The seen case shows quality reconstructions with error evenly distributed throughout the spatial field. The unseen case though has significant error and less coherent reconstructions. Notably, in the sparse unseen case, it can visually be seen that the model utilizes the DT mask and Voronoi mask to reconstruct the ground truth. The greatest error in the unseen reconstruction in Fig.~\ref{fig:unseen_10_optimal_noaa} is in the Atlantic Ocean, and the shape of the error resembles a region of the Voronoi mask. Also note how the further away a point is from a sensor, the lower the prediction value. This can visually be seen in the Indian Ocean. These reconstructions exhibit the dependence a model has on the Voronoi and DT mask inputs under unseen sensor placements.

%%% NOAA Pred Examples:

\clearpage

% 50 Unseen Optimal Sensors Reconstruction

\newpage
\clearpage

% 50 Unseen Baseline Sensors Reconstruction

\newpage
\clearpage

% 10 Seen Optimal Sensors Reconstruction

\newpage
\clearpage

% 10 Unseen Optimal Sensors Reconstruction

\newpage

%%%%%%%%%%%%%%%%%%%%
%%%%% Channel %%%%%
%%%%%%%%%%%%%%%%%%%%
\subsection{Channel Dataset Results}

\begin{table}[b]
\caption{Channel experiment results}
\scriptsize
\begin{center}
\begin{tabular}{l |c c c | c c c c c}
\hline
Experiment & \multicolumn{3}{c|}{Seen Sensor Amounts} & \multicolumn{5}{c}{Unseen Sensor Amounts}\\
\hline
 & 50 & 100 & 200 & 50 & 100 & 150 & 200 & 250 \\
\hline
\hline
Baseline  & 0.597 & 0.437 & 0.308 & 0.591 & 0.440 & 0.362 & 0.314 & 0.276 \\
\hline
Normalized & 0.597 & 0.438 & 0.308 & 0.590 & 0.441 & 0.363 & 0.317 & 0.281 \\
DT mask & 0.596 & 0.437 & 0.309 & 0.589 & 0.441 & 0.363 & 0.316 & 0.279 \\
Normalized \& DT mask & 0.598 & 0.438 & 0.309 & 0.591 & 0.442 & 0.362 & 0.315 & 0.278 \\
Only Voronoi Input & 0.610 & 0.459 & 0.342 & 0.679 & 0.515 & 0.430 & 0.382 & 0.340 \\
\hline
\end{tabular}
\end{center}
\label{channel_results}
\end{table}

Finally we evaluate the reconstruction performance of our methods on the Channel dataset and observe the challenges of the DT mask and normalization. In Table~\ref{channel_results}, we provide a comparison using normalization, a DT information mask, and a combination of both to the baseline method. We also experimented with training a model only using the Voronoi representation as input with no sensor masks. 

Opposed to the Cylinder and NOAA datasets, normalization and the DT mask showed no improvement for the channel dataset. As seen in Table~\ref{channel_results} the ``Baseline'', ``Normalized'', ``DT mask'', and ``Normalized \& DT mask'' experiments exhibited near identical performance for all sensor amounts and for both seen and unseen sensor placements. Though the standard deviations are not reported in the table, paired t-tests confirmed there was no significant difference between the four experiments. While the DT information mask did not improve results, it also did not hinder results. Thus, knowing information about the distance between locations and sensors in non-cyclical and chaotic datasets seems uninformative. With no distinction between these four experiments, we questioned if knowings the sensor locations has any impact at all. Thus, we experimented with training a model only using the Voronoi representation as input with no location or DT masks. As seen in Table~\ref{channel_results}, using no location masks decreased model performance in all cases. Thus, we can conclude that while our proposed data augmentations do not improve reconstruction accuracy, knowing the locations of sensors is crucial.

\subsubsection{Channel Validation Curves}

% \begin{figure}[t]
%     \centering
%     \includegraphics[width = 0.49\linewidth]{images/Val_Curves/channel_validation_curve.png}
% \caption{Validation curves for Channel dataset}
% \label{channel_val_curve}
% \end{figure}

% \begin{figure}
%     \centering
%     \includegraphics[width = 0.49\linewidth]{images/Channel_Images/blurred_channel_val.png}
% \caption{Validation curves for Gaussian blurred Channel datasets}
% \label{blurred_channel_val_curve}
% \end{figure}

\begin{figure}
\centering
\begin{minipage}{.5\textwidth}
    \centering
    \includegraphics[width = \linewidth]{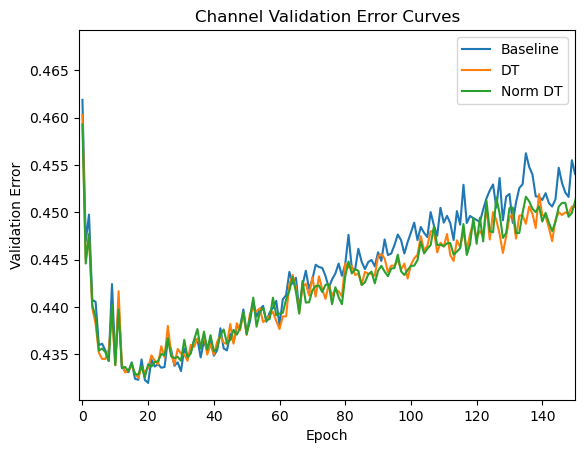}
\caption{Validation curves for Channel dataset}
\label{channel_val_curve}
\end{minipage}%
\begin{minipage}{.5\textwidth}
    \centering
    \includegraphics[width = \linewidth]{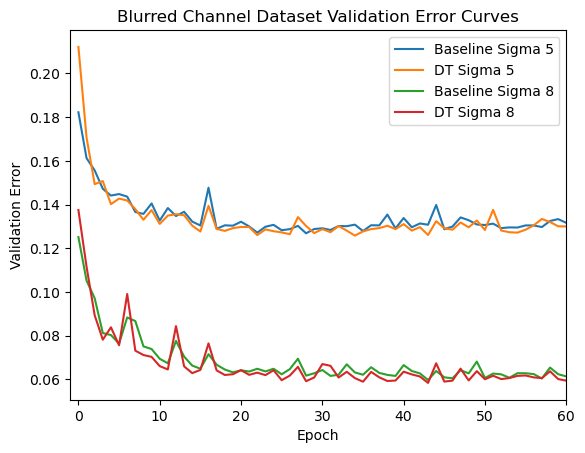}
\caption{Validation curves for Gaussian blurred Channel datasets}
\label{blurred_channel_val_curve}
\end{minipage}
\end{figure}

Analyzing the validation curves for the Channel dataset shows that all models started to overfit in the first 30 epochs. These results highlight the inability for our model to generalize from the training dataset because of the data's chaotic and highly detailed nature. Also note that the validation curves for all our models are near identical, which paints a visual picture of the similar error rates in Table~\ref{channel_results}.

%%%%%%%%%%%%%%%%%%%%%%%%%%%%
%%%% Gaussian Channel Images
%%%%%%%%%%%%%%%%%%%%%%%%%%%%
\begin{figure}[t]
\centering
\begin{subfigure}{.33\linewidth}
    \includegraphics[width = 0.95\linewidth]{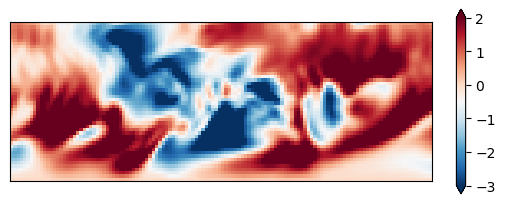}
    \caption{}
    \label{fig_gaussian:no_gauss}
\end{subfigure}% 
\begin{subfigure}{.33\linewidth}
    \includegraphics[width = 0.95\linewidth]{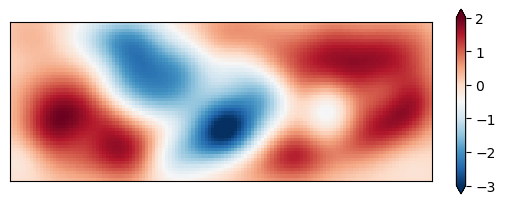}
    \caption{}
    \label{fig_gaussian:5_gauss}
\end{subfigure}% 
\begin{subfigure}{.33\textwidth}
    \includegraphics[width = 0.95\linewidth]{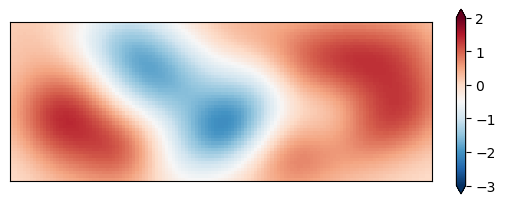}
    \caption{}
    \label{fig_gaussian:8_gauss}
\end{subfigure}
\caption{Gaussian blurred Channel dataset. Examples have a) no blur, b) Gaussian blur with $\sigma = 5$, and c)Gaussian blur with $\sigma = 8$}
\label{gaussian_blur}
\end{figure}

The challenges of our methods on the Channel dataset are attributed to the dataset's non-cyclical and high-frequency data, as seen in Fig.~\ref{fig_gaussian:no_gauss}. We hypothesize that with limited sensors and no underlying cyclical pattern to rely on, this complex detail becomes difficult to predict and reconstruct.

% Possibly mention a high freqeuncy cyclical dataset.
This hypothesis does not address why the DT mask provides no improvement though. While our prior experiments showcased that the DT mask is less effective on datasets that are non-cyclical and have high-frequency, we did not answer which characteristic of the dataset is responsible for the DT mask's inefficiencies. By applying a Gaussian blur over our Channel dataset images, we are able to construct a blurred dataset that has low-frequency yet retains the non-cyclical nature of the Channel dataset. Examples of this dataset with varying blurs can be seen in Fig.~\ref{gaussian_blur}. Reconstruction on a blurred Channel dataset will evaluate whether the non-cyclical nature of a dataset without high frequency detail still presents difficulties for reconstruction on the DT mask.

Validation curves for reconstructing the blurred Channel dataset are presented in Fig.~\ref{blurred_channel_val_curve}. Note that the differences in validation error between models trained on different Gaussian blurs is predictable because blurred images are more homogeneous and easier to predict resulting in lower error rates. The figure depicts that the DT mask does not improve model performance over the baseline model regardless of blur, which highlights the fact that the DT mask is less effective for the channel non-cyclical dataset regardless of if the data has high or low frequency. Furthermore, when comparing to our unblurred validation curves in Fig.~\ref{channel_val_curve} we see the blurred datasets were able to learn and not overfit. This can be attributed to the low-frequency data since reconstructions can be better approximated. An example of this generalization can be seen in the following prediction visualizations. Note how even the DT model in Fig.~\ref{fig:channel_dt_recon} generates a Gaussian blurred looking image highlighting that all of our models cannot reconstruct the highly frequent data.

We propose that non-cyclical datasets cause independence between points in the spatial field. This independence renders metrics such as distance between points a less effective statistic for reconstruction. Thus, adding a DT mask to a model will have little effect since the added information is not helpful for spatially independent points.

% Not sure about this final paragraph
From the lack of reconstruction accuracy, we have shown that the Channel dataset, or non-cyclical and highly detailed datasets, is not as suited for ML-based reconstruction approaches.

\subsubsection{Channel Reconstruction Examples}

\begin{figure}[t]
% Inputs
\begin{subfigure}{.49 \linewidth}
    \centering
    \includegraphics[width = \linewidth]{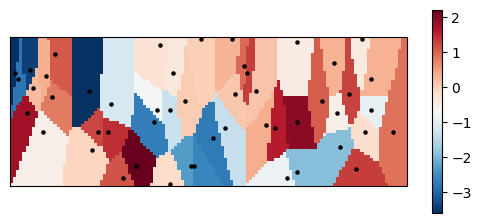}
    \caption{Voronoi}
\end{subfigure}
\begin{subfigure}{.49 \linewidth}
    \centering
    \includegraphics[width = \linewidth]{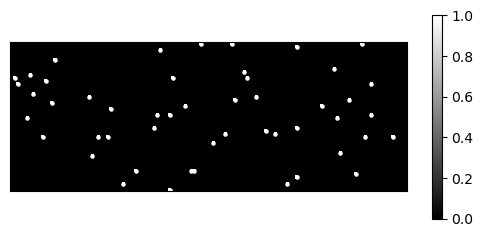}
    \caption{DT Mask}
\end{subfigure}% 

% Prediction
\begin{subfigure}{0.49\linewidth}
    \centering
    \includegraphics[width = \linewidth]{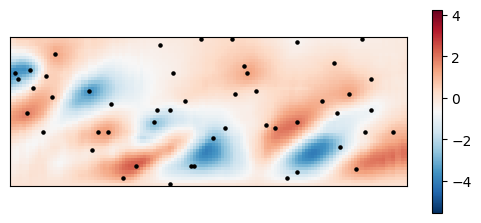}
    \caption{Image Reconstruction (Prediction)}
\end{subfigure}%
\begin{subfigure}{0.49\linewidth}
    \centering
    \includegraphics[width = \linewidth]{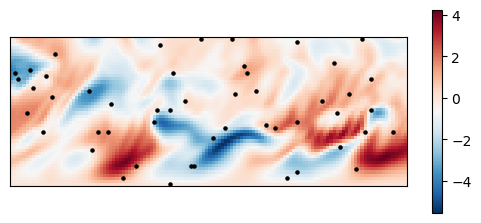}
    \caption{Ground Truth}
\end{subfigure}

% Forth Row of images
\begin{subfigure}{0.49 \linewidth}
    \centering
    \includegraphics[width = \linewidth]{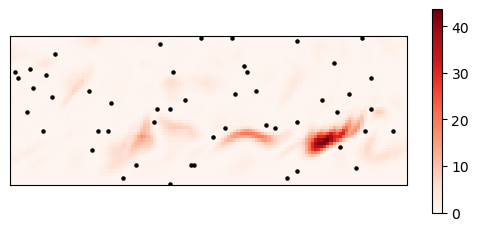}
    \caption{Full Error Map}
\end{subfigure}
\begin{subfigure}{0.49 \linewidth}
    \centering
    \includegraphics[width = \linewidth]{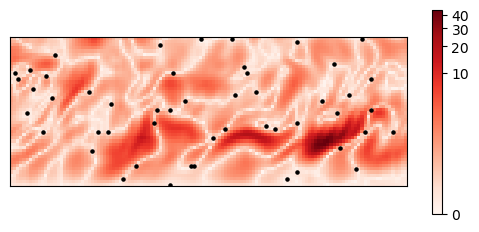}
    \caption{Scaled Error Map}
\end{subfigure}

\caption{Channel dataset reconstruction example for 50 unseen sensors using baseline model}
\label{fig:channel_recon_baseline}
\end{figure}

\begin{figure}[t]
% Inputs
\begin{subfigure}{.49 \linewidth} 
    \centering
    \includegraphics[width = \linewidth]{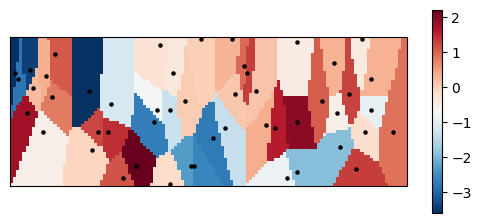}
    \caption{Voronoi}
\end{subfigure}
\begin{subfigure}{.49 \linewidth}
    \centering
    \includegraphics[width = \linewidth]{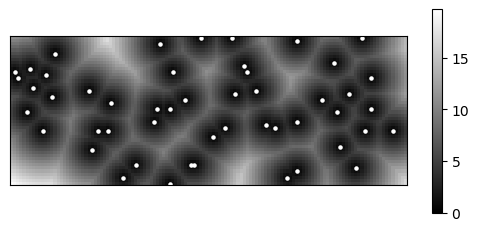}
    \caption{DT Mask}
\end{subfigure}% 

% Prediction
\begin{subfigure}{0.49\linewidth}
    \centering
    \includegraphics[width = \linewidth]{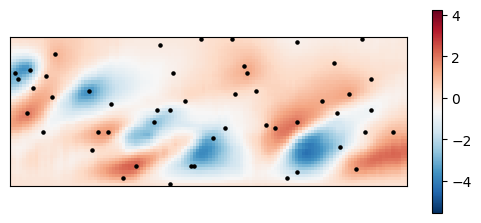}
    \caption{Image Reconstruction (Prediction)}
\end{subfigure} %
\begin{subfigure}{0.49\linewidth}
    \centering
    \includegraphics[width = \linewidth]{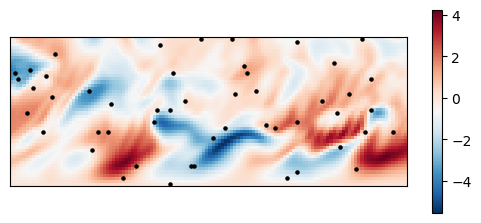}
    \caption{Ground Truth}
\end{subfigure}

% Forth Row of images
\begin{subfigure}{0.49 \linewidth}
    \centering
    \includegraphics[width = \linewidth]{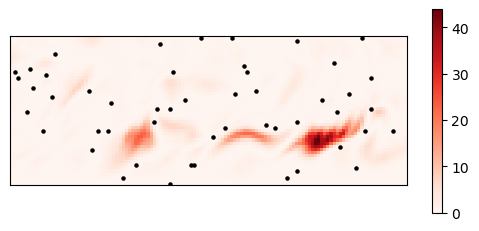}
    \caption{Full Error Map}
\end{subfigure}
\begin{subfigure}{0.49 \linewidth}
    \centering
    \includegraphics[width = \linewidth]{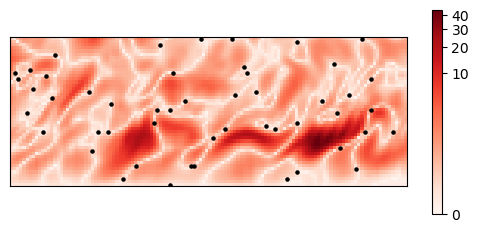}
    \caption{Scaled Error Map}
\end{subfigure}

\caption{Channel dataset reconstruction example for 50 unseen sensors using a DT model}
\label{fig:channel_dt_recon}
\end{figure}

The first two reconstruction examples for the Channel dataset in Fig.~\ref{fig:channel_recon_baseline} and Fig.~\ref{fig:channel_dt_recon} compare the baseline model reconstruction to the DT model reconstruction. The predicted images under both of these models are near identical. Furthermore, their error maps are also indistinguishable. This identical reconstruction shows the fact that adding a DT information mask is not useful for reconstruction for the non-cyclical Channel dataset. Additionally, note how the reconstructions for both models do not capture the complex detail of the ground truth data because of the non-cyclical and chaotic nature of the dataset as described earlier.

\begin{figure}[t]
% Inputs
\begin{subfigure}{\linewidth}
    \centering
    \includegraphics[width = 0.57\linewidth]{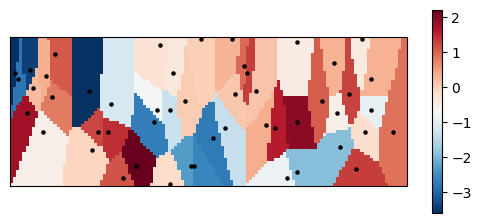}
    \caption{Voronoi}
\end{subfigure}

% Prediction
\begin{subfigure}{0.49\linewidth}
    \centering
    \includegraphics[width = \linewidth]{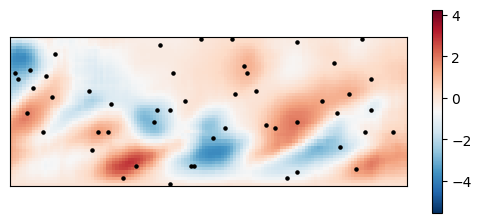}
    \caption{Image Reconstruction (Prediction)}
\end{subfigure} %
\begin{subfigure}{0.49\linewidth}
    \centering
    \includegraphics[width = \linewidth]{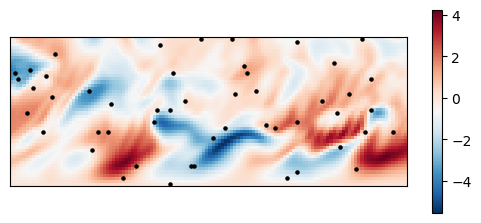}
    \caption{Ground Truth}
\end{subfigure}

% Forth Row of images
\begin{subfigure}{0.49 \linewidth}
    \centering
    \includegraphics[width = \linewidth]{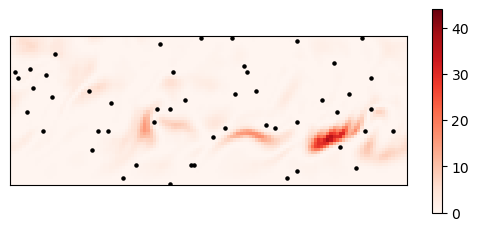}
    \caption{Full Error Map}
\end{subfigure}
\begin{subfigure}{0.49 \linewidth}
    \centering
    \includegraphics[width = \linewidth]{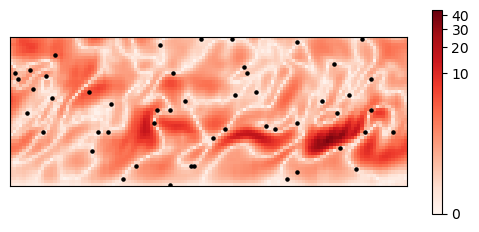}
    \caption{Scaled Error Map}
\end{subfigure}

\caption{Channel dataset reconstruction example for 50 unseen sensors using a single channel}
\label{fig:single_channel_channel}
\end{figure}

The third reconstruction example for the Channel dataset in Fig.~\ref{fig:single_channel_channel} depicts the experiment of reconstructing under only the Voronoi mask. These reconstructions also look near identical to the baseline model, but if analyzed closely there are areas with slightly more error. While not obvious for this example, this reconstruction shows there is a small difference created in removing the sparse mask from the Channel reconstruction model.

\begin{figure}[t]
% Inputs
\begin{subfigure}{.49 \linewidth} 
    \centering
    \includegraphics[width = \linewidth]{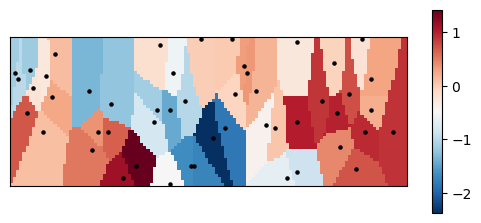}
    \caption{Voronoi}
\end{subfigure}
\begin{subfigure}{.49 \linewidth}
    \centering
    \includegraphics[width = \linewidth]{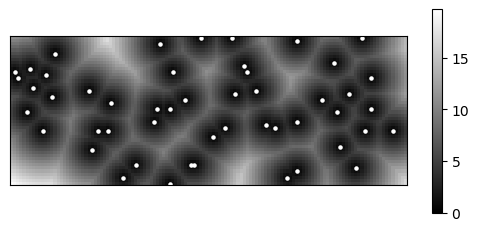}
    \caption{DT Mask}
\end{subfigure}% 

% Prediction
\begin{subfigure}{0.49\linewidth}
    \centering
    \includegraphics[width = \linewidth]{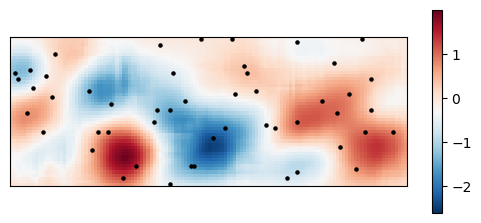}
    \caption{Image Reconstruction (Prediction)}
\end{subfigure} %
\begin{subfigure}{0.49\linewidth}
    \centering
    \includegraphics[width = \linewidth]{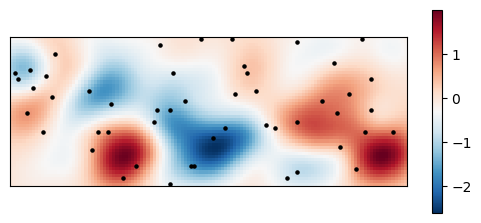}
    \caption{Ground Truth}
\end{subfigure}

% Forth Row of images
\begin{subfigure}{0.49 \linewidth}
    \centering
    \includegraphics[width = \linewidth]{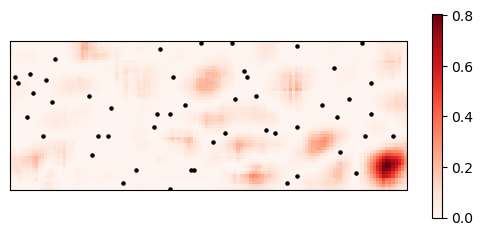}
    \caption{Full Error Map}
\end{subfigure}
\begin{subfigure}{0.49 \linewidth}
    \centering
    \includegraphics[width = \linewidth]{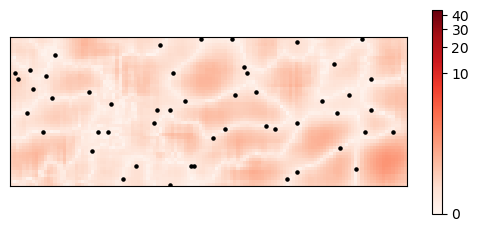}
    \caption{Scaled Error Map}
\end{subfigure}

\caption{Blurred Channel dataset with reconstruction example for 50 unseen sensors using a DT model under $\sigma=5$}
\label{fig:blurred_channel}
\end{figure}

The fourth and final reconstruction example for the Blurred Channel dataset in Fig.~\ref{fig:blurred_channel} depicts the experiment of spatial field reconstruction under a non-cyclical and low-frequency dataset. Opposed to the prior reconstructions, this reconstruction is more accurate as generalization is easier for low frequency data. This graphic visually explains why the blurred datasets had lower error rates and were able to train unlike the unblurred datasets.

%%% Channel Pred Examples:

% 50 seen Sensors Baseline Reconstruction

% 50 seen Sensors DT

% 50 seen Sensors Single Channel Reconstruction

% 50 seen Sensors Blurred Reconstruction

% While the results for reconstruction in \cite{Fukami2021} are better than ours, we were not able to reproduce their results even while using their provided code.

% The channel dataset also presents high frequency data that is chaotic, which is extremely hard to predict with few sensors. When limited to a few sensors the model outputs a very rough approximation that ignores fine detail and tries to capture larger patterns in the model. 

% Since the dataset is acyclical, the model can’t rely on previous data to predict an estimate of the fine detail at adjacent cells. Instead it does an approximation of the data instead.

% Dataset Applications
\section{Application}
\label{application.ch}

The prior three datasets are all derived from the paper \cite{Fukami2021}. While reconstruction of these datasets demonstrate the effectiveness of our approach, it limits the scope of our work to a subset of hand-selected examples. Thus we extrapolated our methods to a final application: reconstruction of sparse climate information in the Antarctic regions.

Accurate climate information of the Antarctic is exceptionally important due to this region's delicate relationship with global warming. The outcomes of climate change on the Antarctic has significant implications for the rest of the global climate system due to the importance of its ice sheets. With increasing temperatures, Antarctic ice sheets are at a greater risk of melting leading to a rise in sea levels and a reduction in the earth's albedo \cite{Desseler2016}. Thus, it is crucial there is accurate and real time monitoring of this region. Though important, the Antarctic has a few amount of weather stations due to the harsh weather conditions, consequently preventing dense analytics. The juxtaposition between the lack of sensors and the importance of measurements presents an obvious application of sparse spatial field reconstruction.

The current solution for solving this problem is through the use of complex reanalysis models. While these methods are successful in creating dense representations of the Antarctic, their latency is quite slow. The current state of the art climate real analysis model led by the European Centre for Medium-Range Weather Forecasts (ECMWF), the leading climate research center in Europe, claims ``real-time'' reconstruction updates \cite{EU_2023}. ``Real-time'' in this case is a three month latency, which is quick for current meteorological standards, but slow in comparisons to other truly real-time metrics \cite{EU_2023}. The lack of speed and true real-time data can be solved by deep learning, as we have seen in the prior examples.

\begin{figure}[t]
\centering
\begin{subfigure}{0.5\linewidth}
    \includegraphics[width = 0.9\linewidth]{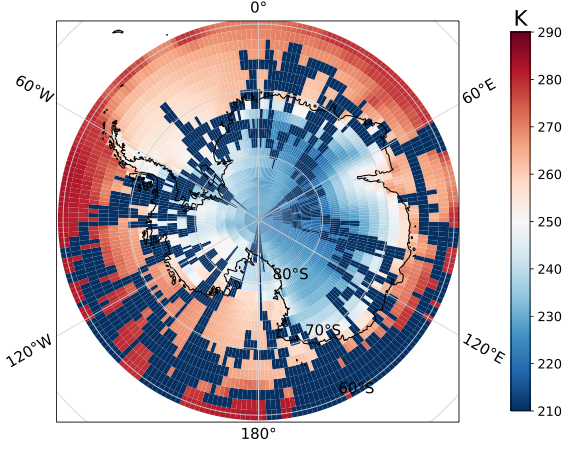}
    \caption{Sparse Mask - 30\% Missing}
    \label{fig:sparse_artic_missing}
\end{subfigure}% 
\begin{subfigure}{0.5\linewidth}
    \includegraphics[width = 0.9\linewidth]{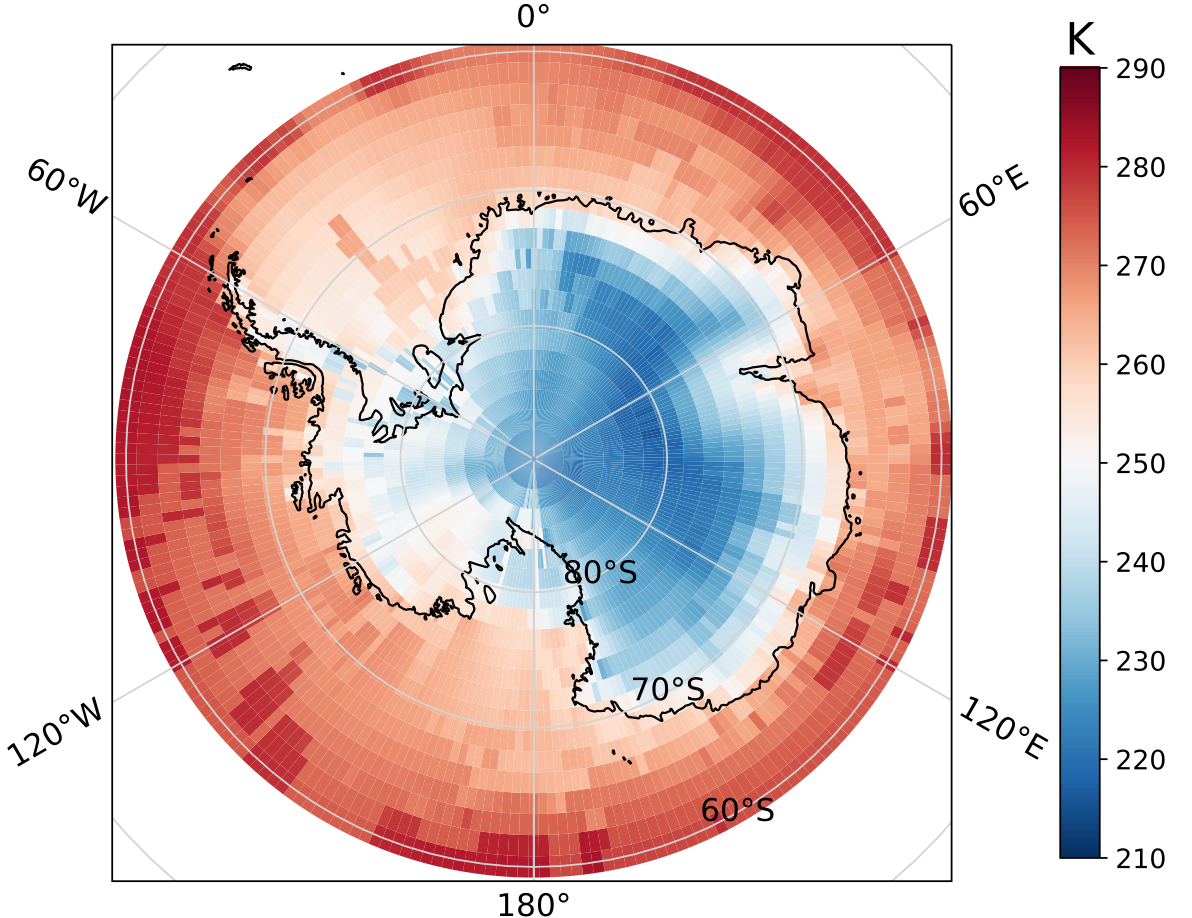}
    \caption{Filled Sparse Mask}
    \label{fig:filled_sparse_artic}
\end{subfigure}
\caption{Example of Input Images of Previous Work}
\label{fig:artic_sparse_mask_ex}
\end{figure}

Deep Learning approaches for reconstructing the Antarctic climate data has been explored by previous research \cite{Yao2023}. In that work, the authors use image in-painting methods to reconstruct the sparse spatial field, but they focus on developing a complex CNN utilizing attention instead of focusing on the input data. The paper does experiment with two input types though. The first is a sparse mask that is filled with the average temperature of the previous 30 years, termed non-physically informed input. An example image can be seen in Fig.~\ref{fig:filled_sparse_artic}. This approach is easy to implement, but the average does not capture the changing temperature distributions based on seasonality, which potentially introduces bias. The second type of input is a sparse mask that is filled with the data from another reanalysis model. This approach is unrealistic in a real time scenario because the model is training with data that is artificially similar to the test data, so we do not explore this input for our reconstructions. Furthermore, this approach is invalid as the other model used to generate the input mask could simply be used as the reconstruction. Thus our approach will compare to a non-physically informed input, which we term the ``Filled Sparse Mask'', because it is a realistic model that could be used in real time. 

Furthermore, this paper experiments with fictitious sparsity, as they create datasets of the Antarctic region that contain 30\%, 50\%, and 70\% missing data to test their reconstruction while the true sparsity exceeds 90\%. An example of their artificially sparse dataset can be seen in Fig.~\ref{fig:sparse_artic_missing}. The paper mainly focuses on the ``ideal'' case of having little sparsity which will never happen in the real-world. Thus, our application will only focus on the true sparsity of the Antarctic.

Finally, we will be testing our results on the simple 8-layer CNN mentioned in Sect.~\ref{sect:model_training} and not the model proposed by \cite{Yao2023} due to the challenges of reproducing their code. Regardless, we show that we can produce lower reconstruction error compared to their best reported error. For this application we seek to improve upon the results of \cite{Yao2023} simply by using our methodologies presented prior.

\subsection{Antarctic Dataset}

\begin{figure}
    \centering
    \includegraphics[width = 0.7\linewidth]{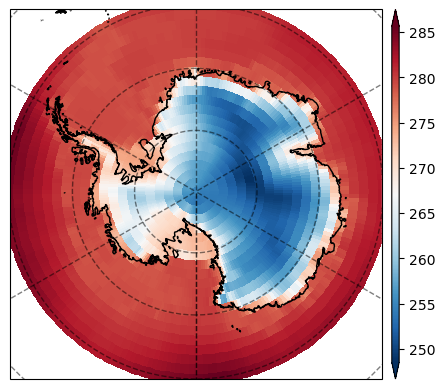}
    \caption{Example Ground Truth Antarctic Surface Temperature in Kelvin}
    \label{fig:6.1}
\end{figure}

The Antarctic dataset used was the monthly average surface temperature from the ERA-Interim reanalysis data created by the European Centre for Medium-Range Weather Forecasts (ECMWF). The dataset spans exactly 40 years, from 1979 to 2018, and focuses on the latitude range $60^{\circ}$S - $90^{\circ}$S (the Antarctic Region) with a spatial resolution of $1.5^{\circ}$ latitude x $1.5^{\circ}$ longitude to only capture the Antarctic Dataset. Numerically, the dataset consists of 480 single-channel images with a resolution of 21x240 pixels representing the surface temperature of the Antarctic in Kelvin. Similar to the Cylinder and NOAA datasets, the Antarctic Dataset is cyclical and repeats climate patterns every 12 images, as 12 images corresponds to a single year. Thus, the Antarctic dataset covers 40 complete cycles.

\subsection{Specific Methodology}

Most methodology used for the application on the Antarctic dataset, such as the Distance Transform and the Voronoi were consistent with prior approaches except for three unique challenges: 1). A non-flat surface for calculating distances, 2). fixed sensors for prediction, and an 3). extrapolated test set.

\subsubsection{Haversine Distance Equation}

\begin{figure}[t]
\centering
\begin{subfigure}{0.49\linewidth}
    \centering
    \includegraphics[width = 0.6\linewidth]{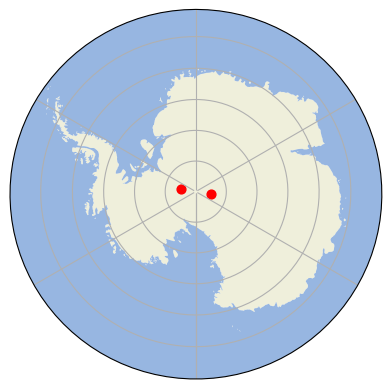}
    \caption{South Pole Stereo Projection}
    \label{fig:stereo_proj}
\end{subfigure}% 
\begin{subfigure}{0.49\linewidth}
    \includegraphics[width = \linewidth]{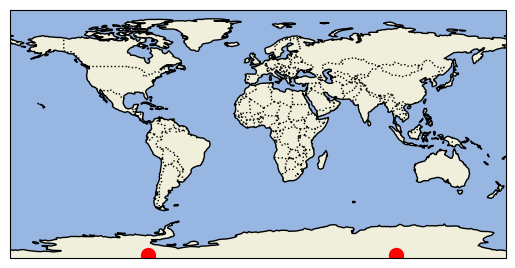}
    \caption{Plate Carr\'ee Projection (Standard Map)}
    \label{fig:platecarree_projection}
\end{subfigure}
\caption{Varying Projections of the Globe }
\label{fig:proj_distortion}
\end{figure}

Maps of the globe are 2-D projections of a 3-D object. Thus any distance calculation done in the 2-D projection space that does not take into account the original 3-D geometry of the globe will inherently contain some error. As discussed earlier for the circular mask in Sect.~\ref{circ_section}, this error is not equally distributed across the 2-D map as varying projections change the distortion in different locations. In Fig.~\ref{fig:stereo_proj}, the Antarctic region was plotted using the South Pole Stereo Projection which visibly has less distortion at the south pole. To compare to other projections, two adjacent stations located at the south pole were plotted as red dots. In Fig.~\ref{fig:platecarree_projection}, the Antarctic region was plotted using the Plate Carr\'ee projection, which is the common map projection used in computational applications due to its perfect Cartesian coordinate system, and the two stations denoted by red dots are now extremely distant from one another. Since our data is a Plate Carr\'ee projection, na\"ive euclidean distance at the poles would cause great distortion in generating the Voronoi or Distance Transform masks. Furthermore, a simple solution such as the circular mask can not solve this problem, making a new approach necessary.

 % These figures showcase the challenge of using a distance metric like Euclidean distance as it computes two different distances depending on the data representation.

The unique solution to this problem is to not calculate the distance between sensors in the 2-D space, but instead in the 3-D space. Thus the Haversine distance formula, which computes the arc distance between two points on a sphere given their latitude and longitude, was utilized. 

The two part equation for the haversine is shown below where $\phi$ denotes latitude, $\lambda$ denotes longitude, and $r$ denotes the radius of the earth.

\begin{equation}\label{eq:haversine}
h = \sin(\frac{\phi_B - \phi_A}{2})^2 + \cos(\phi_A) \cos(\phi_B) \sin(\frac{\lambda_B - \lambda_A}{2})^2
\end{equation}

\begin{equation}\label{eq:dist}
distance = 2r \arcsin(\sqrt{h})
\end{equation}

Using the above equation, we are able to correctly compute the distances between the sensors and generate Voronoi and DT masks that are consistent with the 3-D geometry of the globe.

\subsubsection{Fixed Stations}

The prior datasets in this paper focused on training deep learning models that could handle both sensors that are fixed (seen) and moving (unseen), this is not the case for the Antarctic dataset. All stations in the Antarctic region are fixed and immobile (seen), thus the creation of the sparse masks is slightly different. 

The amount of current sensors is known in the Antarctic and construction of future sensors is minimal or likely known to the subset of scientist who study this region. Thus, our training set sensor locations is likely to mirror that of the test set locations. Upon analyzing the data, it was discovered that sensors frequently go offline, likely due to the harsh conditions of the Antarctic, making training on varying subsets on the initial sensor placements extremely valuable. Thus, the training set of sensors was constructed to randomly sample a subset of sensors from the known locations as a training mask. 

It was found that as little as 41 sensors could be operable to as high as 65 between the years 1998 and 2008, which are the last ten years of the training data. Therefore, multiple sparse masks were constructed for varying sensor amounts between that range and at varying sampling seeds for each sensor amount. Similar to prior methodology, the number of sensors used in a sparse mask comes from the set $n_{sensor, train, antarctic} = \{45,50,55,60,65\}$ with the random seeds used to generate sensor placements coming from the set $n_{seed, train} = \{1,2,3\}.$ Note that there are no random subsets of 65 sensors as this is the maximal sensor set. Thus there are 13 unique sensor masks for training.

\subsubsection{Extrapolation}

The previous three experiments all focused on interpolation, or reconstructing the gaps between training data, but this final dataset is focused on extrapolation. Real time reconstruction focuses on reconstructing future time steps, thus our previous partitioned data split should not be used for testing extrapolation. Instead, we simply used the first 30 years of data for our training set and the last 10 years as our test set to correctly approximate the extrapolation ability of our model.

\subsection{Results}

\begin{table}[t]
\caption{Antarctic application results}
% \scriptsize
\begin{center}
\begin{tabular}{l |c c c |c c c c c}
\hline
Experiment & RMSE\\
\hline
Best Reconstruction in \cite{Yao2023}  &  13.83\\
Filled Sparse Mask  & $5.259 \pm 1.69$ \\
Sparse mask \& Voronoi & $6.970 \pm 2.524$ \\
DT mask \& Voronoi & $6.654 \pm 2.645$ \\
Normalized DT Mask \& Voronoi & $\textbf{4.667} \pm \textbf{1.91}$\\
\hline
\end{tabular}
\end{center}
\label{table_artic}
\end{table}

Results for reconstructing the last 10 years of surface temperature for the Antarctic region are displayed in Table \ref{table_artic} with  Root Mean Squared Error, the metric used for testing in \cite{Yao2023}. The largest reconstruction error is from the best reported reconstruction from the original paper \cite{Yao2023}. Their model and code was attempted to be reproduced but was unable to due to the complexity of their code base. Thus we take their reported error as an estimate of how their model performed on the true reconstruction. Large improvements in reconstruction accuracy were made by using the simple 8-layer model as all other experiments exhibited lower error. Simply by changing the model, there was a decrease in RMSE by roughly 8 Kelvin. This large improvement denotes that using image in-painting models on extremely sparse data is not as effective as a simple CNN. Separating the filled sparse mask into a Voronoi for the sensor readings and a sparse mask for the sensor locations did not improve results as expected though. Adding the DT mask in place of the sparse mask did not improve upon the RMSE of the ``Filled Sparse Mask'', but it marginally improved upon the ``Sparse Mask \& Voronoi''. The lower RMSE of using the ``Filled Sparse Mask'' models is described by its generic prediction that limits error but has little value in reconstruction. This is further discussed in the Antarctic reconstruction examples. Lastly when normalization was applied to the model using the DT mask and Voronoi, we achieved the best reconstruction with the lowest RMSE and a small standard deviation denoting consistent predictions.

\subsubsection{Antarctic Validation Curves}

\begin{figure}[t]
    \centering
    \includegraphics[width = 0.7\linewidth]{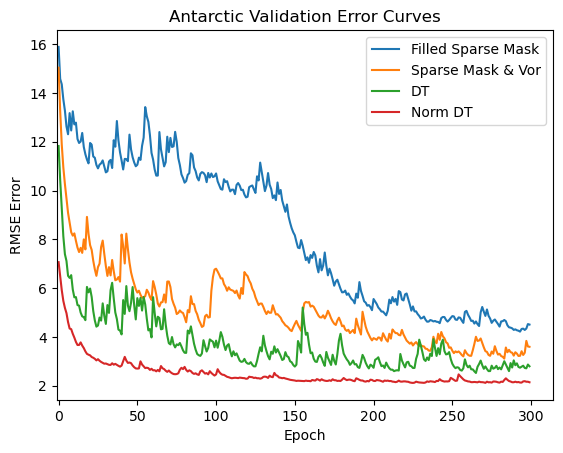}
    \caption{Antarctic Validation Training Curves}
    \label{antarctic_val_curve}
\end{figure}

Similar to earlier experiments, the validation error curves for the Antarctic dataset in Fig.~\ref{antarctic_val_curve} exhibit the benefit a DT mask and normalization has on computational training speed. Note, as modifications were added to the ``Filled Sparse Mask'', such as separating the filled mask into a sparse mask and a Voronoi mask, the training speed increased. Furthermore, each added augmentation decreased the validation error. The model using the ``Filled Sparse Mask'' required 150 epochs to truly start learning while all other models learned quickly. Also note that the ``Filled Sparse Mask'' model had the worst validation error during training, but during the test dataset it performed better than the ``Sparse Mask \& Voronoi'' model as well as the ``Sparse Mask \& DT'' model. While unintuitive, this is described by the default prediction of the average climate mask image that was used to fill in the missing data, which minimizes test and validation error but does not generalize well.

% Weave in the story saying that since the filled sparse mask has been trained on an average of the past information, it will default to this average mask which should generalize to some degree and minize test error. While the test error is low, the reconstructions have less meaning.
% By training on a larger subset of station dropouts, this shift from training and validation set accuracies will drop. 

\subsubsection{Average Error Masks}

\begin{figure}[ht]
\centering
\begin{subfigure}{0.33\linewidth}
    \includegraphics[width = \linewidth]{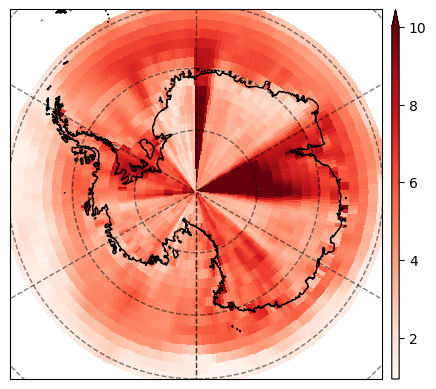}
    \caption{Filled Sparse Mask}
    \label{fig:avg_error_filled}
\end{subfigure}% 
\begin{subfigure}{0.33\linewidth}
    \includegraphics[width = \linewidth]{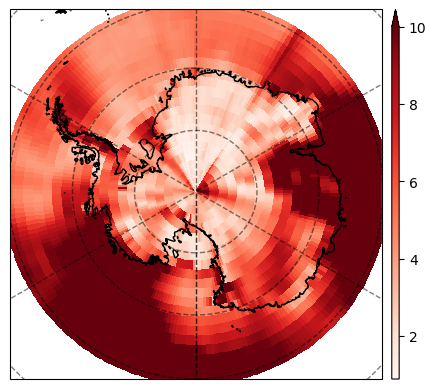}
    \caption{Baseline}
    \label{fig:avg_error_baseline}
\end{subfigure}% 
\begin{subfigure}{0.33\linewidth}
    \includegraphics[width = \linewidth]{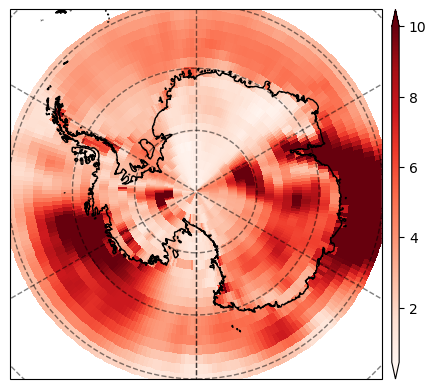}
    \caption{DT Norm}
    \label{fig:avg_error_optimal}
\end{subfigure}
\caption{Antarctic average RMSE images}
\label{fig:arctic_avg_mask}
\end{figure}

The proposed methods in this paper improved reconstruction performance compared to previous methods, but a RMSE of 4.667 Kelvin is still quite large in the context of climate. Thus, the question becomes where is the error occurring most and if this error is acceptable. We answered these questions by generating average error reconstructions across the entire test dataset, pictured in Fig.~\ref{fig:arctic_avg_mask}. 

The ``Filled Sparse Mask'' average error image, while it has a minimal average error overall, generates error in unnatural patterns. The streaking lines of error denote a reconstruction that is not consistent with true climate data. Thus, while the ``Filled Sparse Mask'' reduces error, it does not do it in a physical informed fashion.

The Baseline error mask reconstruction shows that separating the Filled Sparse Mask into a Voronoi and sparse location mask enables the model to reduce the unnatural error patterns and generate more realistic reconstructions. Though the reconstructions seem more consistent, there still is minor streaking error patterns and large error over the oceans. The error regions over the ocean are likely due to the lack of sensor information / estimates in these regions. The Filled Sparse Mask on the other hand is able to reconstruct the ocean as it has the average ocean temperature in the past. Since there are no sensors in the ocean, the Baseline model naturally has more error over the ocean than the Filled Sparse Mask.

Finally, the DT Norm error mask reconstruction shows a cohesive and consistent error map where all the error regions seem natural. It exhibits error in similar locations as the Baseline model but to a lesser degree. To truly know if the DT Norm reconstruction is useful with an average error of 4.667 Kelvin, experts will need to be consulted to see if the current maximal error regions are important to accurately reconstruct. If not our methodologies could be deployed, but if so these error reconstructions are still valuable to assist in choosing future sensor locations in order to minimize error in reconstructions.

\subsubsection{Reconstruction Examples}

\begin{figure}[ht]
% Inputs
\begin{subfigure}{ \linewidth}
    \centering
    \includegraphics[width = 0.4\linewidth]{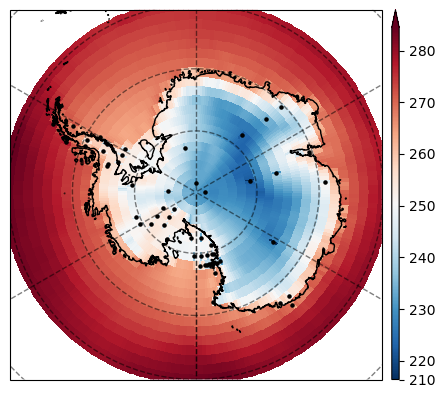}
    \caption{Filled Sparse Mask}
\end{subfigure}% 

% Second Row
\begin{subfigure}{0.49\linewidth}
    \centering
    \includegraphics[width = 0.8\linewidth]{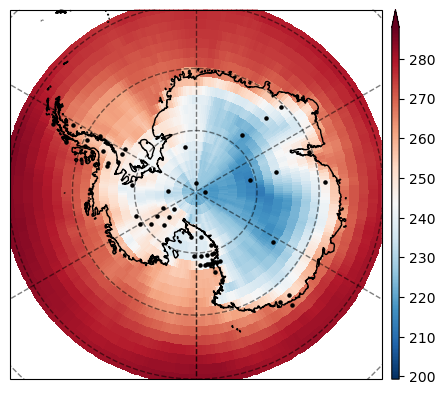}
    \caption{Image Reconstruction (Prediction)}
\end{subfigure}
\begin{subfigure}{0.49\linewidth}
    \centering
    \includegraphics[width = 0.8\linewidth]{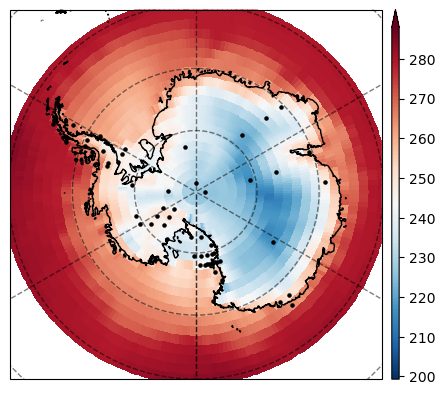}
    \caption{Ground Truth}
\end{subfigure}

% Third Row of images
\begin{subfigure}{0.49 \linewidth}
    \centering
    \includegraphics[width = 0.8\linewidth]{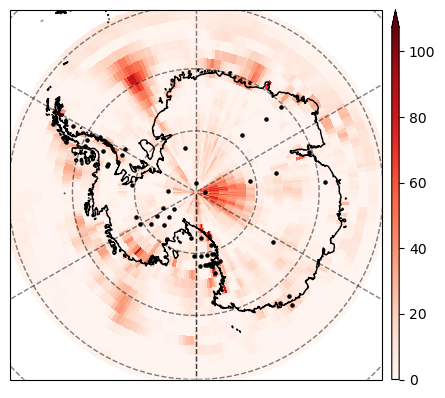}
    \caption{Full Error Map}
\end{subfigure}
\begin{subfigure}{0.49 \linewidth}
    \centering
    \includegraphics[width = 0.8\linewidth]{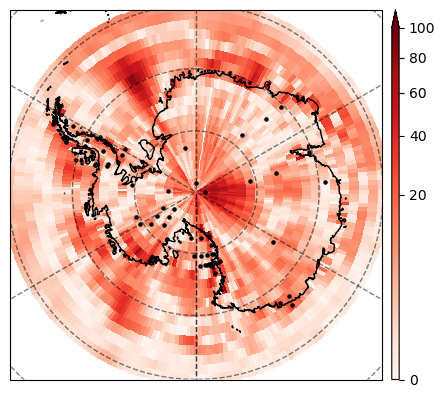}
    \caption{Scaled Error Map}
\end{subfigure}

\caption{Antarctic dataset with reconstruction example using a Filled Sparse Mask Model at the first example in the test dataset}
\label{fig:filled_mask_idx0}
\end{figure}

\begin{figure}[ht]
% Inputs
\begin{subfigure}{.49 \linewidth}
    \centering
    \includegraphics[width = 0.8\linewidth]{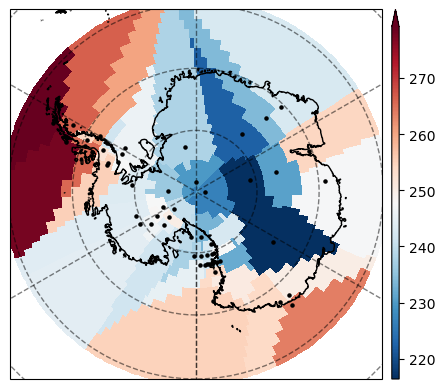}
    \caption{Voronoi}
\end{subfigure}
\begin{subfigure}{.49 \linewidth}
    \centering
    \includegraphics[width = 0.8\linewidth]{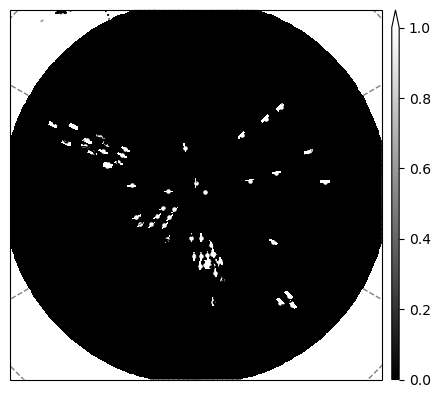}
    \caption{Sparse Mask}
\end{subfigure}% 

% Second Row
\begin{subfigure}{0.49\linewidth}
    \centering
    \includegraphics[width = 0.8\linewidth]{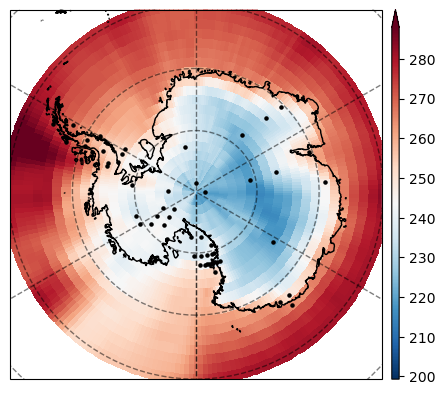}
    \caption{Image Reconstruction (Prediction)}
\end{subfigure}
\begin{subfigure}{0.49\linewidth}
    \centering
    \includegraphics[width = 0.8\linewidth]{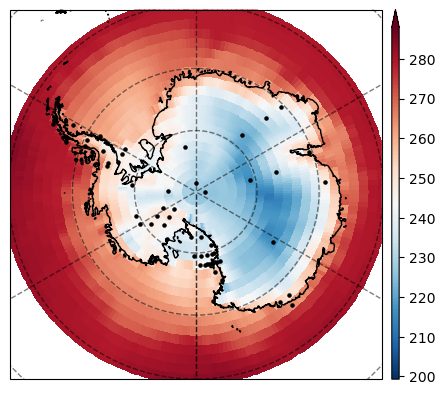}
    \caption{Ground Truth}
\end{subfigure}

% Third Row of images
\begin{subfigure}{0.49 \linewidth}
    \centering
    \includegraphics[width = 0.8\linewidth]{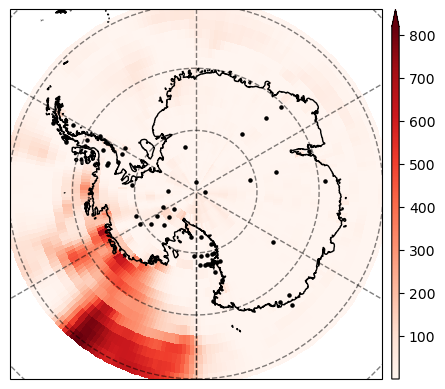}
    \caption{Full Error Map}
\end{subfigure}
\begin{subfigure}{0.49 \linewidth}
    \centering
    \includegraphics[width = 0.8\linewidth]{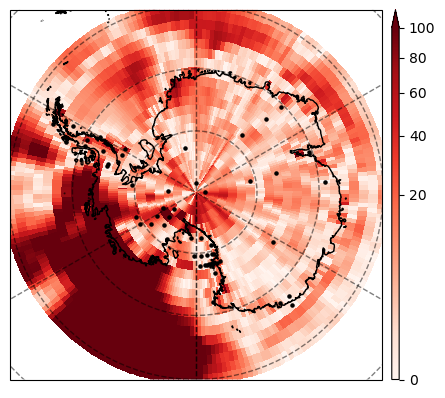}
    \caption{Scaled Error Map}
\end{subfigure}

\caption{Antarctic dataset with reconstruction example using a Baseline Model (Sparse Mask \& Voronoi) at the first example in the test dataset}
\label{fig:baseline_antarctic_recon}
\end{figure}

\begin{figure}[ht]
% Inputs
\begin{subfigure}{.49 \linewidth}
    \centering
    \includegraphics[width = 0.8\linewidth]{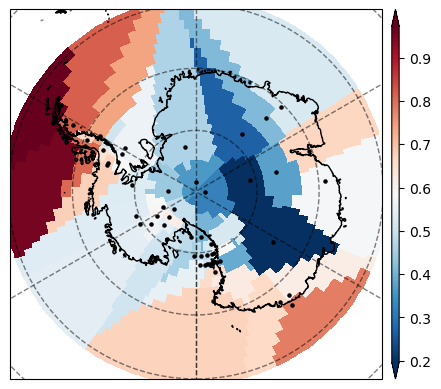}
    \caption{Voronoi}
\end{subfigure}
\begin{subfigure}{.49 \linewidth}
    \centering
    \includegraphics[width = 0.8\linewidth]{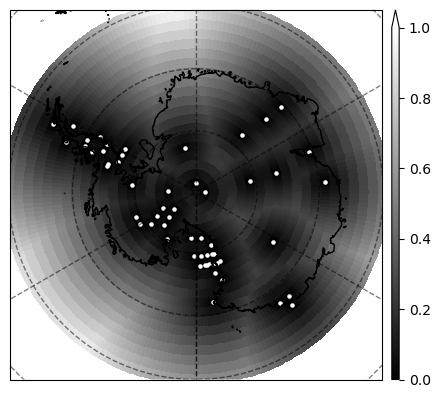}
    \caption{Sparse Mask}
\end{subfigure}% 

% Second Row
\begin{subfigure}{0.49\linewidth}
    \centering
    \includegraphics[width = 0.8\linewidth]{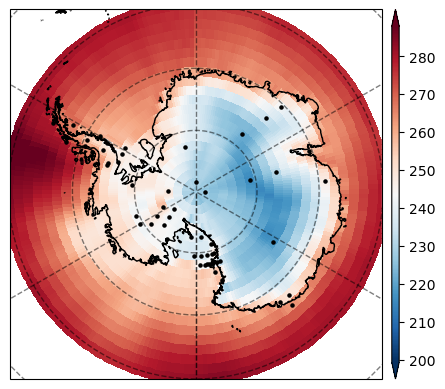}
    \caption{Image Reconstruction (Prediction)}
\end{subfigure}
\begin{subfigure}{0.49\linewidth}
    \centering
    \includegraphics[width = 0.8\linewidth]{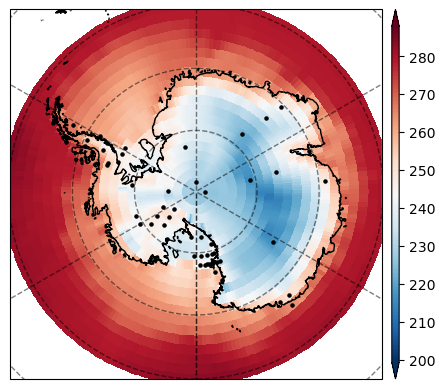}
    \caption{Ground Truth}
\end{subfigure}

% Third Row of images
\begin{subfigure}{0.49 \linewidth}
    \centering
    \includegraphics[width = 0.8\linewidth]{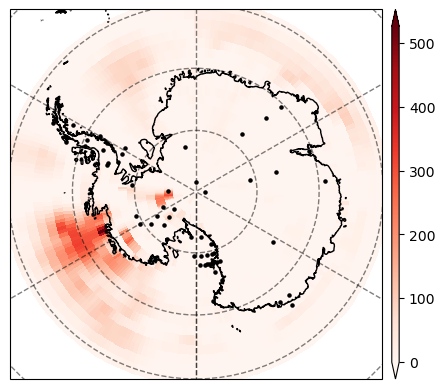}
    \caption{Full Error Map}
\end{subfigure}
\begin{subfigure}{0.49 \linewidth}
    \centering
    \includegraphics[width = 0.8\linewidth]{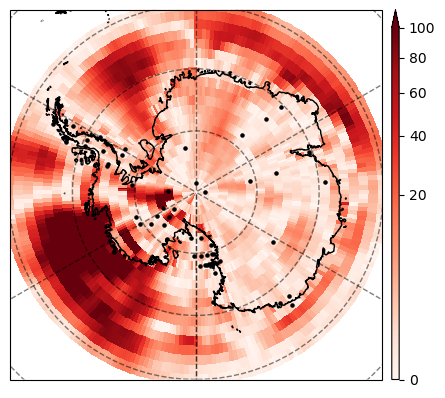}
    \caption{Scaled Error Map}
\end{subfigure}

\caption{Antarctic dataset with reconstruction example using a normalized Model using the DT mask at the first example in the test dataset}
\label{fig:norm_dt_antarctic_recon_idx0}
\end{figure}

The first three reconstructions present the differences between models using the Filled Sparse Mask input from \cite{Yao2023}, the baseline approach from \cite{Fukami2021} using a Sparse Mask \& Voronoi, and our proposed method of normalization with the DT mask. The reconstruction using the Filled Sparse Mask, depicted in Fig.~\ref{fig:filled_mask_idx0}, shows accurate reconstructions. The error mask denotes very little error overall, but this error is non-systematic and irregular as discussed in the prior Average Error Mask section. Note how the Filled Sparse Mask image looks nearly identical to the ground truth before reconstruction, denoting our model might have simply learned an identity function to reduce error. The lack of generalization for the Filled Sparse Mask is explored in the next paragraph. The reconstruction using the Baseline approach is depicted in Fig.~\ref{fig:baseline_antarctic_recon} and shows quality reconstruction on the land mass of the Antarctic but significant error in the ocean. This is attributed to the lack of sensors on the ocean surface. The reconstruction using the Norm DT approach, depicted in Fig.~\ref{fig:norm_dt_antarctic_recon_idx0}, shows a reconstruction similar to the Filled Sparse Mask model reconstruction, expect for more error in the oceans. Note that all errors in the Antarctic region for the Norm DT approach seem natural and consistent as opposed to the streaks and inconsistencies of the Filled Sparse Mask.

\begin{figure}[ht]
% Inputs
\begin{subfigure}{ \linewidth}
    \centering
    \includegraphics[width = 0.4\linewidth]{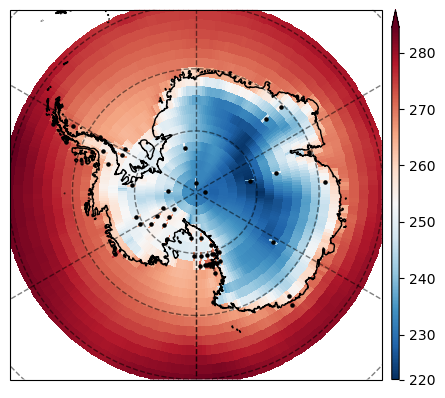}
    \caption{Filled Sparse Mask}
\end{subfigure}% 

% Second Row
\begin{subfigure}{0.49\linewidth}
    \centering
    \includegraphics[width = 0.8\linewidth]{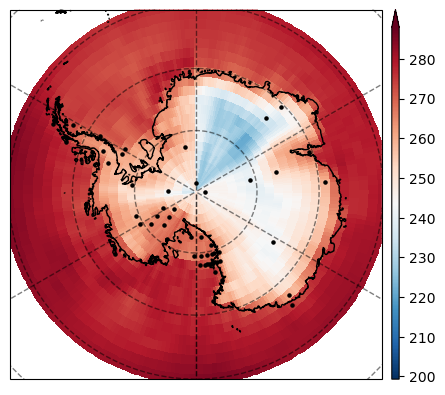}
    \caption{Image Reconstruction (Prediction)}
\end{subfigure}
\begin{subfigure}{0.49\linewidth}
    \centering
    \includegraphics[width = 0.8\linewidth]{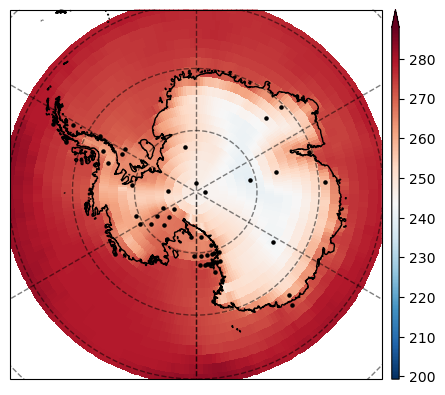}
    \caption{Ground Truth}
\end{subfigure}

% Third Row of images
\begin{subfigure}{0.49 \linewidth}
    \centering
    \includegraphics[width = 0.8\linewidth]{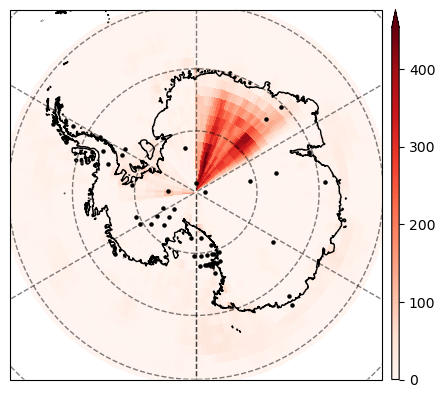}
    \caption{Full Error Map}
\end{subfigure}
\begin{subfigure}{0.49 \linewidth}
    \centering
    \includegraphics[width = 0.8\linewidth]{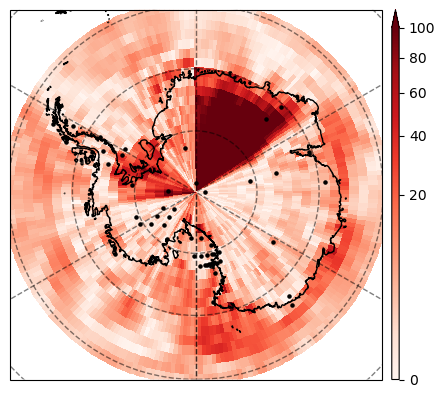}
    \caption{Scaled Error Map}
\end{subfigure}

\caption{Antarctic dataset with reconstruction example using a Filled Sparse Mask Model at the sixth example in the test dataset}
\label{fig:filled_mask_idx6}
\end{figure}

\begin{figure}[ht]
% Inputs
\begin{subfigure}{.49 \linewidth}
    \centering
    \includegraphics[width = 0.8\linewidth]{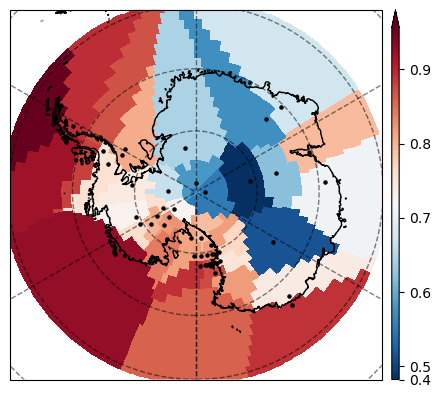}
    \caption{Voronoi}
\end{subfigure}
\begin{subfigure}{.49 \linewidth}
    \centering
    \includegraphics[width = 0.8\linewidth]{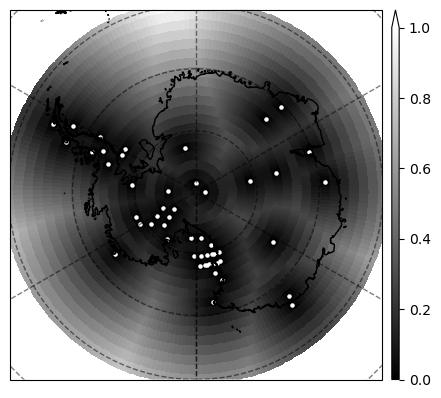}
    \caption{Sparse Mask}
\end{subfigure}% 

% Second Row
\begin{subfigure}{0.49\linewidth}
    \centering
    \includegraphics[width = 0.8\linewidth]{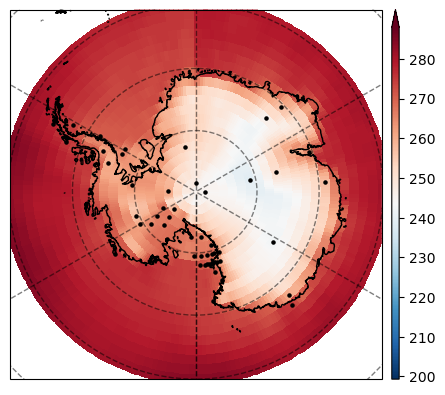}
    \caption{Image Reconstruction (Prediction)}
\end{subfigure}
\begin{subfigure}{0.49\linewidth}
    \centering
    \includegraphics[width = 0.8\linewidth]{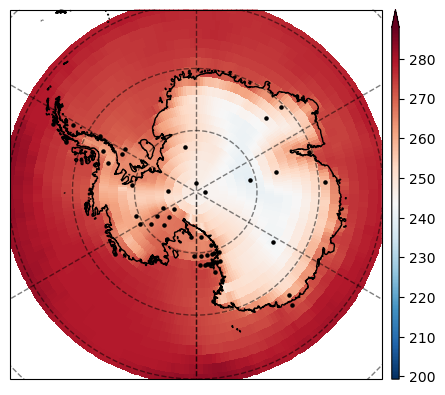}
    \caption{Ground Truth}
\end{subfigure}

% Third Row of images
\begin{subfigure}{0.49 \linewidth}
    \centering
    \includegraphics[width = 0.8\linewidth]{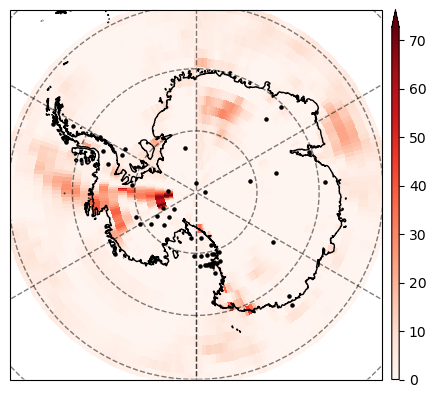}
    \caption{Full Error Map}
\end{subfigure}
\begin{subfigure}{0.49 \linewidth}
    \centering
    \includegraphics[width = 0.8\linewidth]{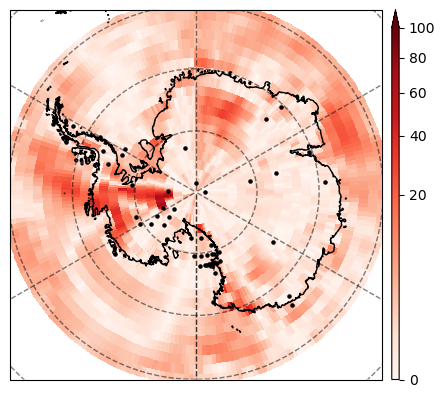}
    \caption{Scaled Error Map}
\end{subfigure}

\caption{Antarctic dataset with reconstruction example using a normalized Model using the DT mask at the sixth example in the test dataset}
\label{fig:norm_dt_antarctic_recon_idx6}
\end{figure}

The last two reconstructions compare the Filled Sparse Mask approach to the Norm DT approach on another test sample that is in the summer. These reconstructions show the generalizing capabilities of our approach and the short comings of the Filled Sparse Mask approach. In Fig.~\ref{fig:filled_mask_idx6}, we see that Filled Sparse Mask image no longer resembles the test dataset causing peculiar errors, such as the patch of extreme cold, in the reconstructed image. Thus, the Filled Sparse Mask approach can not generalize for different seasons. This is not the case for the Norm DT approach, pictured in Fig.~\ref{fig:norm_dt_antarctic_recon_idx6}, which has an accurate reconstruction of the ground truth image. These experiments highlight the fact that the Norm DT approach allows a model to learn true generalization and reconstruction ability as opposed to the identity-esque model of the Filled Sparse Mask approach.

% Filled Mask Model idx 0

% Baseline Model Recon 

% Optimal Model Recon idx 0

% Filled Mask Model idx 6

% Optimal Model Recon idx 6

% Figure 5.12: Channel dataset reconstruction example for 50 seen sensors using base-
% line model

\subsubsection{Application Conclusion}

Reconstructing the Antarctic dataset to an accurate degree in real time can greatly improve the latency of reanalysis models to date. This application showed that our methods presented in this paper could be used to reduce the RMSE error of current reconstruction methodology by several Kelvin. While our reconstructions improved upon past research, the reconstructions still have large error compared to the real ground truth image, thus there is more work needed before this methodology can be fully trusted and deployed in the Antarctic surface temperature reconstruction domain.

% Dataset Conclusions
\section{Conclusion}
\label{conclusion.ch}

\subsection{Summary}

In this work, we proposed improvements to an existing CNN based sparse data spatial field reconstruction method. We proposed applying the Distance Transform to the sparse sensor location mask images, normalizing the input and their associated ground truth images, and separating out domain knowledge to channels such as the land mask. We demonstrated our approach on three datasets from various scientific domains and observed that our approach can provide lower reconstruction error rates for the two-dimensional Cylindrical wake and the NOAA sea surface temperature datasets. However, we observed that our approach did not lead to performance gains for the Channel dataset, but rather discovered that ML-based reconstruction is not suitable for this problem entirely. From our experiments, we choose to reconstruct an Antarctic surface temperature dataset to demonstrate the effective application of our methods and achieved better reconstructions than previous work.

\subsection{Contributions and Significance}

This paper introduces novel data augmentations that can be used to improve sparse spatial field reconstruction. Specifically:

\begin{itemize}
  \item \textbf{The Distance Transform Mask:} An input image to a Deep Learning model that denotes the distance of a location to the nearest sensor. This input image introduces information regarding the spatial relationship between points in a spatial field allowing the Deep Learning model to understand that locations far away from sensors have less information than locations close to sensors.
  
  \item \textbf{The Land Mask:} An input image that denotes the locations in a map that are meant to be masked out or have no real values in the reconstructed image. Previous methods combined the Land Mask with other input data, which in turn introduces ambiguity into the data. 

  \item \textbf{Image Normalization:} A simple normalization that bounds the values of all input images, the Distance Transform Mask, Voronoi Mask, Ground Truth, etc, to a range of [-1,1]. Previous methods used no normalization, resulting in slower reconstruction speeds.
  
\end{itemize}

 Utilizing, the above data augmentations resulted in a 2-6x increase in training speed compared to previous Deep-Learning based reconstruction methods as well as a decrease in reconstruction error.

 By improving sparse spatial field reconstruction, we are presenting a method that geoscientists can use to accurately reconstruct historical data or reduce the latency of current reanalysis models. This increase of  available data is useful to understand the effects of climate change both historically and in the future. Furthermore, the field of fluid dynamics will benefit from more accurate reconstructions by being able to make better decisions, like active flow control, under limited information.

\subsection{Future Work}

The methodology in this paper presents the opportunity for ample future work such as: reconstructing historical datasets, incorporating time-series information, and experimenting with model architectures.

\begin{itemize}
  \item \textbf{Reconstructing Historical Data:} 
  
  The introduction to this paper stated there are two types of applications of Deep Learning to improve sparse spatial field reconstruction. The first is developing Deep Learning models to reduce the data latency in reanalysis models, which was the sole focus of this work. The second, which was not explored, is to develop Deep Learning models that are used to reconstruct sparse historical climate data given dense current data, effectively replacing current reconstruction simulations. 
    
  Future work could focus on exploring the usefulness of our methods in the second application, or restoring historical data. Such examples could include reconstructing the sub-seasurface temperature or salinity from the Argo datasets. Our methodology can be very beneficial to this domain of problems as it has sparse historical data that would be valuable to reconstruct and dense current data to train on. 
  
  \item \textbf{Introducing Time:} 

   This paper treated every image of the dataset as individual and independent instances, while in reality these sequences have dependence between each other, like a Markov chain. Utilizing the dependence the current state of an image has on the previous state, we can potentially improve our model and improve the reconstruction of the next frame. Even if adjacent frames to predict from are not available, introducing a time component into the reconstruction process could be very valuable.

  \item \textbf{Model Architectures:}

This paper used a simple 8-layer CNN for reconstruction leaving opportunity to improve performance by experimenting with more complex state of the art models. Some possible models to experiment with are residual networks and networks that utilize attention such as \cite{Yao2023}.
  
\end{itemize}

\bibliographystyle{unsrt}  
\bibliography{bibfile}

\end{document}